\documentclass[runningheads]{llncs}
\usepackage{graphicx}

\usepackage{tikz}
\usepackage{comment}
\usepackage{amsmath,amssymb} 

\usepackage[accsupp]{axessibility}  

\usepackage{booktabs}
\usepackage{bm}

\usepackage{graphbox}
\usepackage{paralist}

\usepackage{makecell}
\usepackage{tabularx}
\usepackage{multirow}
\usepackage{pifont}
\usepackage{caption}
\usepackage{subcaption}
 \usepackage[T1]{fontenc}
\usepackage[font=small,labelfont=bf,tableposition=top]{caption}
\DeclareCaptionLabelFormat{andtable}{\tablename~\thetable  \&  #1~#2}

\usepackage{algorithm}
\usepackage{setspace}
\usepackage[noend]{algpseudocode}

\usepackage{color}
\usepackage{xcolor,colortbl}
\definecolor{MyGreen}{cmyk}{100, 0, 100, 0}
\usepackage[pagebackref,breaklinks,colorlinks,citecolor=MyGreen]{hyperref}

\DeclareMathOperator*{\argmax}{arg\,max}

\def\Minus{\texttt{-}}
\def\Equal{\texttt{=}}

\newcommand{\etal}{\textit{et al.}}

\begin{document}
\pagestyle{headings}
\mainmatter
\def\ECCVSubNumber{5937}  
\title{CADyQ: Content-Aware Dynamic Quantization \\for Image Super-Resolution}
\titlerunning{CADyQ: Content-Aware Dynamic Quantization for Image Super-Resolution}

\author{Cheeun Hong\inst{1} \and
Sungyong Baik\inst{3} \and
Heewon Kim\inst{1} \and\\
Seungjun Nah\inst{1,4} \and
Kyoung Mu Lee\inst{1,2}
}

\authorrunning{C. Hong \etal}

\institute{
Dept. of ECE \& ASRI, \email{\{cheeun914, ghimhw, kyoungmu\}@snu.ac.kr} \and
IPAI, Seoul National University \and
Dept. of Data Science, Hanyang University \email{dsybaik@hanyang.ac.kr} \and
NVIDIA \email{seungjun.nah@gmail.com} 
}

\maketitle
\begin{abstract}
Despite breakthrough advances in image super-resolution (SR) with convolutional neural networks (CNNs), SR has yet to enjoy ubiquitous applications due to the high computational complexity of SR networks. Quantization is one of the promising approaches to solve this problem. 
However, existing methods fail to quantize SR models with a bit-width lower than 8 bits, suffering from severe accuracy loss due to fixed bit-width quantization applied everywhere. In this work, to achieve high average bit-reduction with less accuracy loss, we propose a novel \textbf{C}ontent-\textbf{A}ware \textbf{Dy}namic \textbf{Q}uantization (CADyQ) method for SR networks that allocates optimal bits to local regions and layers adaptively based on the local contents of an input image. 
To this end, a trainable bit selector module is introduced to determine the proper bit-width and quantization level for each layer and a given local image patch. 
This module is governed by the quantization sensitivity that is estimated by using both the average magnitude of image gradient of the patch and the standard deviation of the input feature of the layer. 
The proposed quantization pipeline has been tested on various SR networks and evaluated on several standard benchmarks extensively. Significant reduction in computational complexity and the elevated restoration accuracy clearly demonstrate the effectiveness of the proposed CADyQ framework for SR.
Codes are available at \hyperlink{https://github.com/Cheeun/CADyQ}{https://github.com/Cheeun/CADyQ}.
\end{abstract}
\section{Introduction}
\label{sec:introduction}

\begin{figure}[t]
    \centering
    \includegraphics[width=0.98\textwidth]{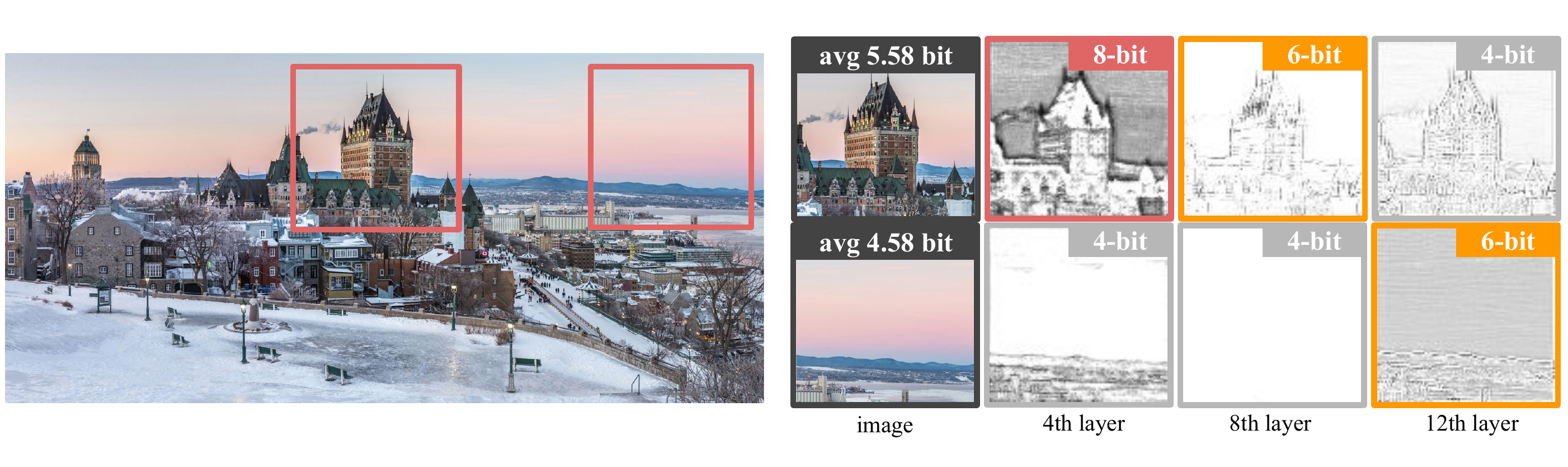}
    \caption{
        \textbf{The dynamic bit-width allocation by CADyQ.} 
        Examples from CADyQ applied to a recent SR network, CARN~\cite{ahn2018fast}.
        Our framework demonstrates a dynamic bit-width allocation per patch and layer with a minimal PSNR drop (<0.05 dB).
        Higher bit-widths are allocated to features containing more structural information or contours
    }
    \label{fig:intro-first}
\end{figure}

Image super-resolution (SR) is a fundamental low-level computer vision problem that aims to restore the high-resolution (HR) image from its corresponding low-resolution (LR) image.
Owing to the remarkable success in deep learning approaches~\cite{dong2015image,kim2016accurate,lim2017enhanced,son2021srwarp,zhang2018rcan,zhang2018residual}, high-fidelity images could be obtained using state-of-the-art super-resolution networks. 
Such modern deep learning models, however, rely on advanced architectures with high computational costs, thereby limiting their applications, especially in resource-limited environments.

Quantization is one of the promising approaches for reducing the computational complexity of neural networks. 
In particular, network quantization has greatly reduced computation loads without a significant accuracy loss, especially for high-level vision tasks (\textit{e.g.}, classification)~\cite{choi2018pact,hou2018loss,zhou2016dorefa}. 
Recently, there have also been attempts to quantize SR networks, either by learning parameters for the binarization of each convolution weight~\cite{ma2019efficient} or by learning a quantization range for each layer~\cite{Li2020pams}. 
However, unlike quantizing high-level vision networks, quantizing SR networks to bit-width lower than 8 bit while maintaining the performance remains a challenging problem~\cite{ignatov2021real}.

\begin{figure}[t]
    \begin{subfigure}{0.5\textwidth}
        \centering
        \includegraphics[width=\textwidth]{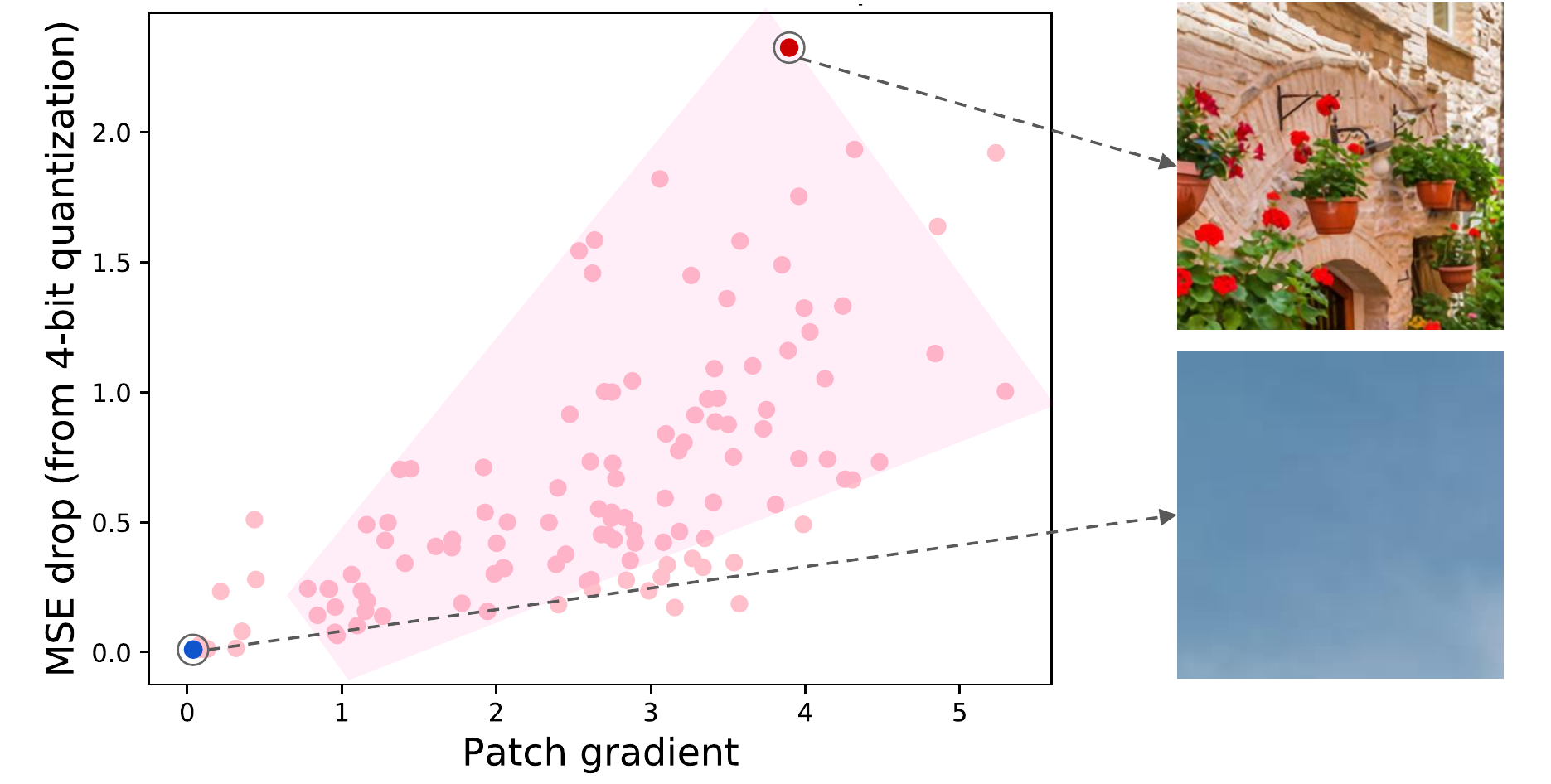}
        \caption{Quantization sensitivity v.s. patch image gradient \label{fig:method-observation-a}}
    \end{subfigure} 
    \hfill
    \begin{subfigure}{0.5\textwidth}
        \centering
        \includegraphics[width=\textwidth]{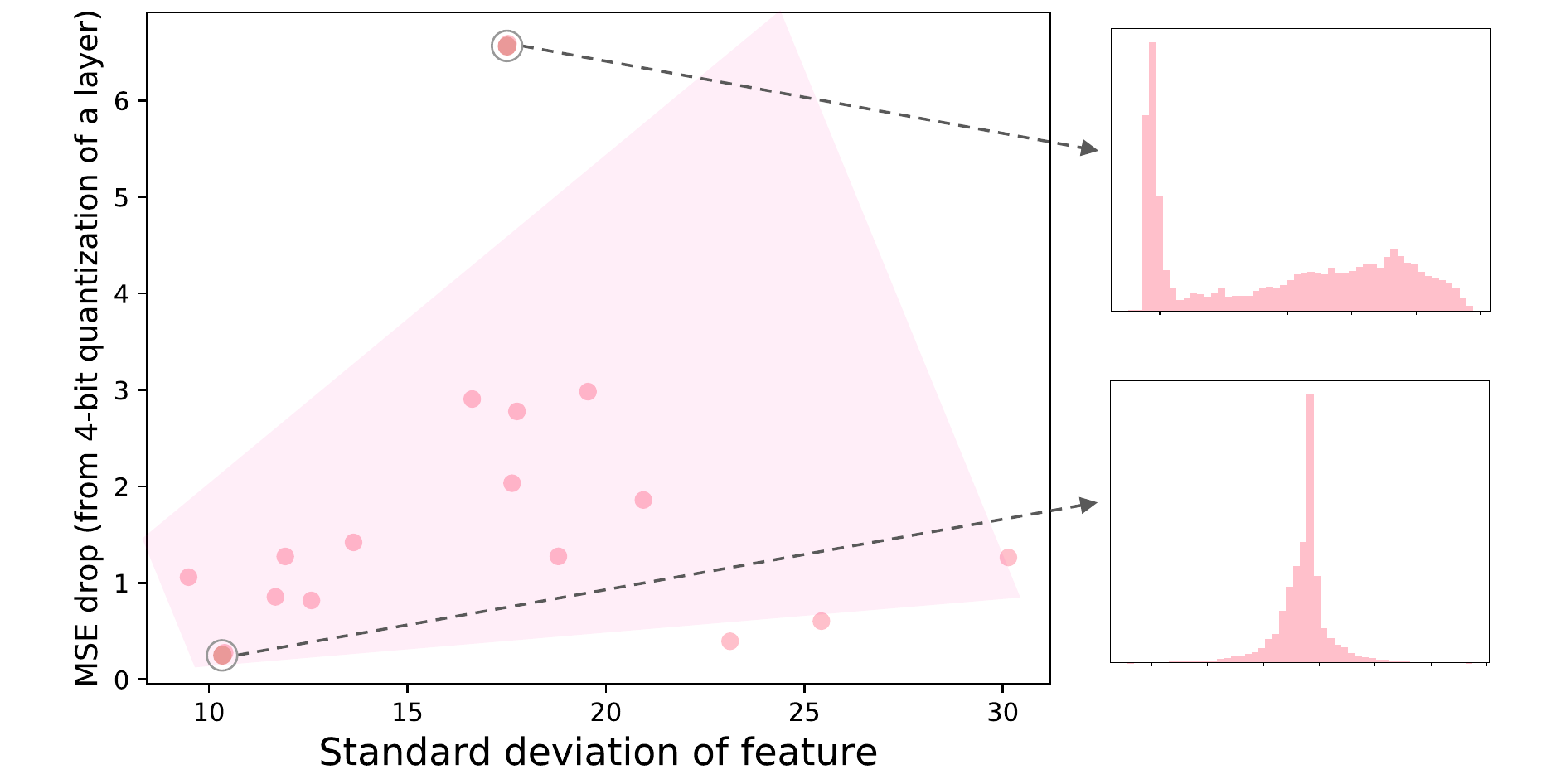}
        \caption{Quantization sensitivity v.s. standard deviation of layer features\label{fig:method-observation-b}}
    \end{subfigure}
    \caption{ 
        \textbf{The motivation of our framework: the different quantization sensitivity per layer and per patch.} Quantization sensitivity is measured with restoration performance (\textit{e.g.}, mean-square error (MSE) between the output image and ground-truth HR image) degradation due to quantization. We observe \textbf{(a)} a correlation between the average magnitude of image gradient~\cite{fattal2007image} and the quantization sensitivity of each patch. Patches with complex (simple) structures exhibit high (low) image gradient magnitude and suffer more (less) MSE drop from quantization. Also, we notice a \textbf{(b)} strong correlation between feature standard deviation and the quantization sensitivity of each layer feature for the given patch. Layers with high (low) feature standard deviation bring more (less) MSE drop from quantizing the given layer.
    }
    \label{fig:method-observation}
\end{figure}

As a key to the above issue, we find that the existing methods do not consider the image structure and locality information, employing a quantized network with fixed bit for all regions of given input images. 
This leads to processing image regions of less structural information with unnecessarily high bits. 
In this work, we observe that different local regions (\textit{i.e.}, patches of a certain size) exhibit different amounts of SR performance degradation from quantization, as illustrated in Fig.~\ref{fig:method-observation-a}. 
In particular, patches with complex structures or contents tend to suffer more from performance degradation than patches with simple contents. 
Furthermore, we also observe that the quantization sensitivity varies among layers, even for the same patch, as illustrated in Fig.~\ref{fig:method-observation-b}. 
The observations suggest that different patches and layers require different precision and thus different bit-widths, providing motivations for a dynamic patch-and-layer-wise bit-width quantization.

Therefore, we propose a new quantization pipeline, dubbed \textbf{C}ontent-\textbf{A}ware \textbf{Dy}namic \textbf{Q}uantization (CADyQ), that dynamically selects a quantization bit-width for each convolution layer based on the quantization sensitivity of its input contents (i.e., each patch and layer feature), as demonstrated in Fig.~\ref{fig:intro-first}. 
However, the direct measurement of the quantization sensitivity is unsuitable as it requires ground-truth high-resolution images to measure the performance degradation. 
In order to estimate the quantization sensitivity, the proposed pipeline employs the average gradient magnitude of the input patch and the standard deviation of the layer feature, based on the observed correlation in Fig.~\ref{fig:method-observation}. 
Then, we introduce a lightweight bit selector that employs a linear layer conditioned on the estimated quantization sensitivity, to determine the bit-width of the feature for each input patch and layer.

Furthermore, a new regularization loss function is introduced to facilitate the bit-width selection process. 
The proposed loss function penalizes the bit selector if a high (low) bit is selected for features with a small (large) quantization sensitivity. 
This leads the bit selector to reserve more computation resources for features that are more critical to the restoration performance, while minimizing the resources for features with less impact on the performance.

The experimental results demonstrate the outstanding performance of the proposed quantization mechanism across various SR networks, underlining the effectiveness and importance of selecting a different bit-width for each patch and layer. 
Overall, our contributions can be summarized as follows:
\begin{compactitem}[$\bullet$]
    \item For the first time, we observe that the sensitivity of restoration accuracy to low-bit quantization varies across different local image regions and the SR network layers.
    \item Accordingly, we present a new quantization framework CADyQ that quantizes SR networks with a different bit-width for each patch and layer, by adding a lightweight bit selector module that is conditioned on the estimated quantization sensitivity.
    \item A novel regularization loss term is introduced to encourage the proposed framework to find a better balance between the computational complexity and overall restoration performance.
\end{compactitem}

\begin{figure}[t]      
    \centering
    \includegraphics[width=1.0\textwidth]{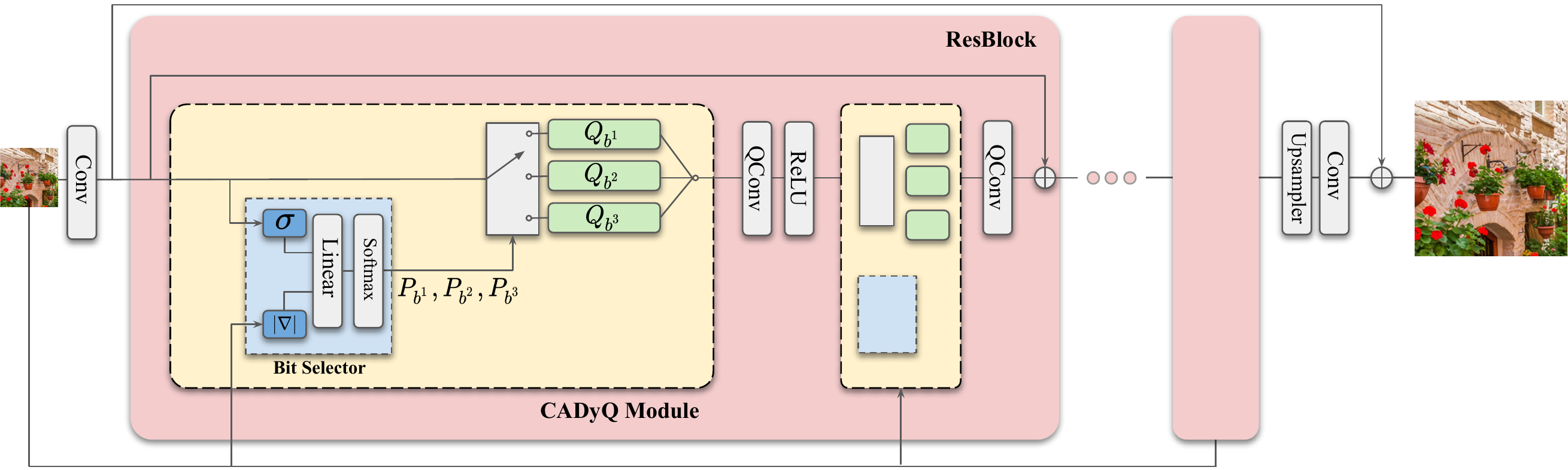}
    \caption{
        \textbf{The overview of the proposed quantization framework CADyQ for SR network}, which we illustrate with a residual block based backbone.
        For each given patch and each layer, our CADyQ module introduces a light-weight bit selector that dynamically selects the bit-width and its corresponding quantization function $Q_{b^{k}}$ (Eq.~\eqref{eqn:method-quantmodule-pams}) among the candidate quantization functions with distinct bit-widths.
        The bit selector is conditioned on the estimated quantization sensitivity (the average gradient magnitude ${|\nabla{}|}$ of the given patch and the standard deviation $\sigma$ of the layer feature).
        Qconv denotes the convolution layer of the quantized features and weights
    }
    \label{fig:method-overview}
\end{figure}
\section{Related Works}
\label{sec:related_works}
\noindent
\textbf{Super-Resolution Neural Networks.} 
Convolutional neural network (CNN) based approaches~\cite{ledig2017photo,lim2017enhanced} have greatly improved the performance of image super-resolution (SR) methods, however, with heavy computational resources.
The massive computations of SR networks have limited the application on real-world mobile devices, spurring the recent interest in lightweight SR networks.
Since then, new lightweight architectures have been investigated~\cite{dong2016srcnn,hui2019imdn,hui2018idn,zhang2018rcan} or searched~\cite{chu2021fast,kim2019fine,li2020dhp,li2021heterogeneity,Oh_2022_CVPR}.
Recently, a few works have introduced adaptive SR networks that aim to achieve efficient inference for a given input~\cite{kong2021classsr,liu2020deep,wang2021exploring,xie2021fadn,yu2021path}. 
These methods mostly focus on reducing the network depth or the number of channels that still rely on heavy floating-point operations, while our focus is to lower the precision of floating-point operations with network quantization.

\noindent
\textbf{Neural Network Quantization.}
Network quantization provides an alternative approach for making networks efficient by mapping 32-bit floating point values of feature maps and weights to lower bit values~\cite{cai2017deep,choi2018pact,esser2019learned,jung2019learning,lee2021network,zhou2016dorefa,zhuang2018towards}. 
Few recent works have attempted to allocate different bit-widths for different layers~\cite{cai2020rethinking,dong2019hawq,habi2020hmq,jin2020adabits,lou2019autoq,wang2019haq,yang2020fracbits}.
However, these approaches target high-level tasks and thus do not consider the distinct local regions that we observe to play a key role in obtaining an efficient network for super-resolution.

\noindent
\textbf{Quantized Super-Resolution Models.}
In contrast to high-level vision tasks, super-resolution poses different challenges due to inherently high accuracy sensitivity to quantization~\cite{ignatov2021real,ma2019efficient,xin2020binarized}.
A few works have attempted to recover the accuracy by modifying the network architecture~\cite{ayazoglu2021extremely,jiang2021training,xin2020binarized}. 
However, the methods are applicable to specific models and thus not generalizable to other SR architectures.
For a general quantization method for SR networks, 
PAMS~\cite{Li2020pams} learns the quantization intervals of different layers to adapt to vastly distinct distributions in the features of SR networks (due to the absence of BN layers), and DAQ~\cite{hong2022daq} further achieves ultra-low bit quantization on SR by utilizing different quantization function parameters for each feature channel.
Furthermore, Wang~\etal~\cite{Wang2021fully} has proposed quantizing features from all layers and skip connections of SR networks. 
Considering the varying degree of quantization sensitivity inside the network, Liu~\etal~\cite{liu2021super} manually allocated a bit-width for each stage of a network.
However, these works apply a fixed bit-width either throughout different input images~\cite{liu2021super} or both images and network layers~\cite{Wang2021fully}.
In contrast, we observe that the quantization sensitivity varies throughout the network layers and images.
Thus, we propose a new quantization framework that dynamically selects the appropriate bit-width for each layer feature based on its quantization sensitivity of each content (\textit{i.e.}, patch and layer feature).
\section{Proposed Method}
\label{sec:proposed_method}
\subsection{Preliminaries}\label{subsec:preliminaries}
Generally, to replace the majority of floating-point operations with lower-bit operations in CNNs, the input feature and weight of each convolutional layer are respectively quantized~\cite{cai2017deep,choi2018pact,jung2019learning}.
Given an input feature of the $j$-th convolutional layer $\bm{x}^j\in\mathbb{R}^{N\times C\times H\times W}$, where $B, C, H,$ and $W$ denote the mini-batch size, number of channels, height, and width of the feature, a quantization function $Q_{b}(\cdot)$ quantizes the feature $\bm{x}^j$ into its low-bit counterpart $\bm{x}^j_q$ of bit-width $b$:
\begin{equation}\label{eqn:method-quantmodule-pre}
    \bm{x}^j_q \equiv Q_{b}(\bm{x}^j) = \lfloor {\text{clamp}(\bm{x}^j, a)} \cdot\frac{s(b)}{a}\rceil \cdot \frac{a}{{s(b)}}.
\end{equation} 
$\bm{x}^j$ is first truncated with $\text{clamp}(\cdot,a)$ into the range of $[-a, a]$, and then scaled to $[-1, 1]$ with the scale parameter $a$.
Then, $\bm{x}^j$ is scaled to the integer range of the given bit-width $b$, $[-s(b), s(b)]$ where $s(b)\Equal{2^{b\Minus1}\Minus1}$.
Consequently, the features in integer range are then rounded to integer values with $\lfloor\cdot\rceil$, and then rescaled back to range $[-a, a]$.
For quantizing features of SR networks, scale parameter $a$ is either implemented with a learnable parameter~\cite{Li2020pams} or a moving average of batch-wise max values~\cite{Wang2021fully}.
At the output of the ReLU layers, since the values are non-negative, the output values are truncated into the range of $[0, a]$ and then scaled to integer range $[0, s(b)]$ with $s(b)\Equal{2^{b}\Minus1}$.
Similarly, a weight of the $j$-th convolutional layer $\bm{w}^j\in\mathbb{R}^{C\times C_{out}\times F\times F}$ is quantized to $\bm{w}^j_q$:
\begin{equation}\label{eqn:method-quantmodule-pre2}
    \bm{w}^j_q \equiv Q_{b}(\bm{w}^j) = \lfloor {\text{clamp}(\bm{w}^j, a^j_w)} \cdot\frac{s(b)}{a^j_w}\rceil \cdot \frac{a^j_w}{{s(b)}},
\end{equation}
where $C$ and $C_{out}$ are the number of input and output channels, $F$ is the kernel size of the convolution filter, and $a^j_w$ is the scale parameter for the corresponding weight $\bm{w}^j$.
In quantization for SR networks, the weight scale parameter $a^j_w$ is often determined simply by $a^j_w=\text{max}(|\bm{w}^j|)$~\cite{Li2020pams}.

\subsection{Motivation}
\label{subsec:motivation}
Previous SR quantization works~\cite{Li2020pams,Wang2021fully} have quantized the network with a fixed bit-width $b$, as formulated in Eq.~\eqref{eqn:method-quantmodule-pre}.
However, our observations in Fig.~\ref{fig:method-observation} hint the disadvantages of a fixed bit-width quantization in SR. 
In particular, different patches and network layers exhibit different degrees of quantization sensitivity (\textit{i.e.}, SR performance drop from a fixed-bit quantization).
As such, we aim to dynamically assign bit-widths for the features based on the quantization sensitivity of \textit{contents} (\textit{i.e.}, the input patch \textit{and} the layer), thereby naming our proposed framework \textbf{C}ontents-\textbf{A}ware \textbf{Dy}namic \textbf{Q}uantization (CADyQ).

\subsection{Proposed Quantization Module (CADyQ)}
\label{subsec:quantization-module}
In our proposed framework, each convolutional layer has a quantization module, which in turn consists of $K$ bit-width quantization function candidates, one of which is selected by a bit selector, as illustrated in Fig.~\ref{fig:method-overview}.
\\

\noindent\textbf{Dynamic Feature Quantization.}
To dynamically quantize features with a different bit-width for each $i$-th patch and $j$-th layer, a single quantization function will be selected in the CADyQ module among $K$ number of candidate quantization functions with distinct bit-widths.
Each quantization function $Q_{b^{k}_{i,j}}(\cdot)$ of bit-width ${b^{k}_{i,j}}$ $(k\Equal1,...,K)$ will, when selected, quantize the feature of the $i$-th patch and $j$-th layer $\bm{x}^j_i$ with 
\begin{equation}\label{eqn:method-quantmodule-pams}
    Q_{b^k_{i,j}}(\bm{x}^j_i) = \lfloor \text{clamp}(\bm{x}^j_i, a_k)\cdot\frac{s({b^{k}_{i,j}})}{a_k} \rceil\cdot\frac{a_k}{s({b^{k}_{i,j}})},
\end{equation} 
where $s(b_k)\Equal{2^{b_k\Minus1}\Minus1}$ is the integer range of the bit-width $b_k$ and $a_k$ is the scale parameter.
Once the bit-width is selected to be $b^{k^*}_{i,j}$ for $i$-th input patch and $j$-th layer, the resulting quantized counterpart of $\bm{x}_i^j$ is
\begin{equation*}\label{eqn:method-quantmodule-out}
    \bm{x}_{i,q}^j \equiv Q_{b^{k^*}_{i,j}}(\bm{x}_{i}^j),
\end{equation*} 
where $Q_{b^{k^*}_{i,j}}$ is the quantization function for $\bm{x}^j_{i}$, corresponding to the selected bit-width ${b^{k^*}_{i,j}}$. 
Note that we simply use a linear symmetric quantization function and a learnable scale parameter $a_k$ for each quantization function, as in~\cite{Li2020pams}.
\\

\noindent\textbf{Bit-Width Selection.}
To facilitate the bit-width selection, we use a lightweight bit selector that assigns a probability to each bit-width. 
Then, the bit-width with the highest probability is selected:
\begin{equation}
    \label{eqn:method-bitloss4}
    {b^{k^*}_{i,j}} = 
    \begin{cases}
    {\argmax_{b^{k}_{i,j}}  P_{b^{k}_{i,j}}(\bm{x}^j_{i}) } & \text{forward}, 
    \\
    \sum_{k=1}^K b^{k}_{i,j}\cdot P_{b^{k}_{i,j}}({\bm{x}^j_i}) & \text{backward},
    \end{cases}
\end{equation} 
where $P_{b^{k}_{i,j}}$ is the probability assigned to the bit-width $b^{k}_{i,j}$ and its corresponding quantization function and $\sum_k{P_{b^{k}_{i,j}}}=1$.
We desire our bit selector network to predict a high probability to a high bit-width for features that have high quantization sensitivity (high accuracy drop from quantization) and a low bit-width for features with low quantization sensitivity.
However, it is infeasible to directly measure the quantization sensitivity of each feature without access to a ground-truth HR patch for the given input LR patch. 
Therefore, upon the correlations observed in Fig.~\ref{fig:method-observation}, we estimate the quantization sensitivity for each layer and the given input patch with the average magnitude of image gradient~\cite{fattal2007image} of a patch and the standard deviation of a feature. 
Conditioned on the average gradient magnitude of a patch and the standard deviation of a feature, our bit selector assigns the probability to each bit-width candidate for $\bm{x}_i^j$, the feature of the $j$-th layer and input patch $I_i$:
\begin{equation}\label{eqn:method-bitmodule-prob}
     P_{b^{k}_{i,j}}(\bm{x}^j_i) =  {\frac{\text{exp} ( f( \sigma(\bm{x}^j_i),|\nabla{I_i}| ) )} {\sum_{k=1}^K {\text{exp} ( f( \sigma(\bm{x}^j_i),|\nabla{I_i}| ) ) }} },
\end{equation} 
where $\sigma(\bm{x}^j_i)\in\mathbb{R}^{C}$ measures the channel-wise standard deviation and $|\nabla{I_i}|\in\mathbb{R}^{2}$ measures the average magnitude of the image gradients from the patch $I_i$ in horizontal and vertical directions~\cite{fattal2007image}.
We concatenate the two metrics and then pass it through a fully connected layer $f : \mathbb{R}^{C+2}\rightarrow \mathbb{R}^{K}$, based on the observed positive correlation between the measured quantization sensitivity and the feature standard deviation or the average gradient magnitude of each patch.
While we make observations on the correlation using the layer-wise standard deviation for the clarity, the bit selector is conditioned on the channel-wise standard deviation, which is observed to have more fine-grained information, as discussed in the supplementary document. 
\\

\noindent\textbf{Backpropagation.}
Selecting a quantization function of the max probability, however, is a discrete non-differentiable process and cannot be optimized end-to-end.
Hence, we employ the straight-through estimator~\cite{bengio2013estimating} to make the process differentiable.
The discrete bit-width selection is replaced with its differentiable approximation, where each candidate bit-width is weighted by the probability distribution predicted by the bit selector (Eq.~\eqref{eqn:method-bitmodule-prob}), during backpropagation:
\begin{equation*}
    \label{eqn:method-bitmodule-softout}
    \bm{x}^j_{i,q} = 
    \begin{cases}
    Q_{b^{k^*}_{i,j}}(\bm{x}^j_{i}) & \text{forward}, 
    \\
    \sum_{k=1}^K Q_{b^{k}_{i,j}}({\bm{x}^j_i})\cdot P_{b^{k}_{i,j}}({\bm{x}^j_i}) & \text{backward}.
    \end{cases}
\end{equation*}

\noindent\textbf{Weight Quantization.}
Weights are quantized with a fixed bit-width, as in Eq.~\eqref{eqn:method-quantmodule-pre2}.
While weights can be also quantized dynamically, we focus on the dynamic quantization of input features, motivated by the observations of the correlations between the quantization sensitivity and local image contents (\textit{i.e.}, patches and layer features) in Fig.~\ref{fig:method-observation}.

\subsection{Bit Loss}
\label{subsec:loss}
Previous works~\cite{Li2020pams,Wang2021fully} have focused on optimizing the performance of a quantized network with a fixed bit-width, by using a pixel-wise L1 loss and knowledge distillation loss~\cite{hinton2015distilling} with the original unquantized network.
On the other hand, we aim to find an efficient quantization for the given feature dynamically.
Hence, we need a regularization loss term to strike a balance between the restoration performance and the quantization rate.
Directed by a similar goal, few neural architecture search (NAS)-based mixed-precision quantization approaches~\cite{cai2020rethinking,yang2020fracbits} utilize {bit regularization loss} to optimize the computational resources of the quantized network.
The typical bit regularization loss penalizes the total number of operations weighted by its bit-width of the currently selected network:
\begin{equation}\label{eqn:method-bitloss-c}
    \mathcal{L}_{b} = \sum_{j=1}^{M} \sum_{i=1}^{N} b^{k^*}_{i,j}\cdot {\text{OPs}(\bm{x}_{i}^j)},
\end{equation} 
where $N$ is the batch size; $M$ is the number of quantized layers in the network; $b^{k^*}_{i,j}$ is the selected bit-width corresponding to the feature of $i$-th patch and $j$-th layer according to Eq.~\eqref{eqn:method-bitloss4}; and ${\text{OPs}}(\cdot)$ is the number of operations for convoluting the given feature.
However, this standard {bit regularization loss} equally penalizes the quantization modules of different layers when each layer can have a different impact on the overall performance after quantization, as observed in Fig.~\ref{fig:method-observation-b}.

To achieve a better trade-off between the computational cost and restoration performance, the bit-widths of quantization modules with a larger impact on performance should be penalized less than those of the quantization modules with less impact.
As a result, the layers with greater impact on the overall performance will have higher bit-width assigned.
To this end, we modify the bit regularization loss by weighting each selected bit-width with the probability estimated by our bit selector, which is conditioned on the estimated quantization sensitivity (and thus estimated impact on the overall performance).
Given feature $\bm{x}_i^j$ of the $j$-th layer for the $i$-th patch, 
our weighted {bit regularization loss} is
\begin{equation} \label{eqn:method-bitloss5}
    \mathcal{L}_{wb} = \sum_{j=1}^{M} \sum_{i=1}^{N} \frac{b^{k^*}_{i,j}}{\sum_k{{b^{k}_{i,j}}\cdot \text{sg}[P_{b^{k}_{i,j}}(\bm{x}_{i}^j)] }} \cdot {\text{OPs}}(\bm{x}_{i}^j),
\end{equation} 
where $\text{sg}[\cdot]$ denotes stop gradient operation. 
Specifically, the denominator represents the expected bit-width during training, while the numerator represents the selected bit-width.
When the expected bit-width is smaller than the selected bit-width, it results in a larger regularization term and hence stronger penalization.
This penalization enforces a lower bit-width assignment on the feature estimated to be less sensitive to quantization.
For example, when the probability distribution from the bit selector network for each bit-width 4, 6, and 8 is [0.1, 0.1, 0.8], the quantization module should be regularized less than [0.2, 0.2, 0.6]. 
The feature that corresponds to the former probability distribution can be considered to be more vulnerable to the performance drop from quantization.
On the other hand, when the expected bit-width is larger than the selected bit-width, a larger expected bit-width is regularized less.
This enables the bit selector to select a higher bit-width for more quantization-sensitive features.

Then, our final objective function becomes
\begin{equation}  \label{eqn:method-loss}
    \mathcal{L} = w_1 \mathcal{L}_1 + w_\text{reg} \mathcal{L}_\text{reg} + w_\text{kd}\mathcal{L}_\text{kd} + w_\text{kdf}\mathcal{L}_\text{kdf},
\end{equation}
where $w_1, w_\text{reg}, w_\text{kd}, w_\text{kdf}$ are the weights to balance different loss terms, respectively; $\mathcal{L}_1$ is the pixel-wise L1 loss between the output image and the ground truth; $\mathcal{L}_\text{reg}$ is a {bit regularization} loss ($\mathcal{L}_\text{wb}$ in our case); $\mathcal{L}_\text{kd}$ is the knowledge distillation loss on the last output feature using 8-bit quantized model as the teacher; $\mathcal{L}_\text{kdf}$ is the knowledge distillation loss on output feature of each layer using the same 8-bit teacher.
As for the knowledge distillation, a teacher network has the same architecture backbone (\textit{i.e.}, no bit selection module included) as the student SR network.
A teacher network is pre-trained with uniform 8-bit weights with the activations quantized via PAMS~\cite{Li2020pams}.
\section{Experiments}
\label{sec:experiments}
The proposed quantization framework CADyQ is evaluated on various SR networks to validate its effectiveness and flexibility.
We first describe our experimental settings (Sec.~\ref{subsec:setting}) and evaluate our method on various SR networks (Sec.~\ref{subsec:comparison}). 
We then present detailed ablation experiments to analyze each main attribute of our framework (Sec.~\ref{subsec:ablation}): namely, layer-wise/patch-wise quantization, quantization sensitivity estimation, and the proposed weighted bit loss.

\subsection{Implementation Details} \label{subsec:setting}
\begin{table}[t]
\centering
\caption{ \textbf{Quantitative comparisons on various SR networks:} 
IDN~\cite{hui2018idn}, EDSR-baseline~\cite{lim2017enhanced}, SRResNet~\cite{ledig2017photo}, and CARN~\cite{ahn2018fast} of scale 4.
The average feature quantization rate of the feature extraction stage (FQR), PSNR, and SSIM are reported.
The results demonstrate the efficiency of the proposed method that manages to reduce FQR while maintaining or improving PSNR/SSIM
}
\label{tab:exp-main}
\resizebox{\textwidth}{!}{
    \begin{tabular}{l rcc rcc rcc}
        \toprule
        \multirow{2}{*}{Model} & 
        \multicolumn{3}{c}{Urban100} & \multicolumn{3}{c}{Test2K} & \multicolumn{3}{c}{Test4K} \\
        \cmidrule(lr){2-4} \cmidrule(lr){5-7} \cmidrule(lr){8-10} & 
        FQR$_\downarrow$ & PSNR$_\uparrow$ & SSIM$_\uparrow$ & 
        FQR$_\downarrow$ & PSNR$_\uparrow$ & SSIM$_\uparrow$ & 
        FQR$_\downarrow$ & PSNR$_\uparrow$ & SSIM$_\uparrow$\\
        \midrule
        IDN~\cite{hui2018idn}&32.00&25.42&0.763&32.00&27.48&0.774&32.00&28.54&0.806\\
        IDN-PAMS~\cite{Li2020pams}&8.00&25.56&0.768&8.00&27.53&0.775&8.00&28.59&0.807\\
        IDN-DAQ~\cite{hong2022daq}&4.00&24.46&0.718&4.00&26.98&0.750&4.00&27.94&0.782\\ 
        \rowcolor{orange!25}
        IDN-CADyQ (Ours)&5.78&25.65&0.771& 5.16&27.54&0.776& 5.03&28.61&0.808\\
        \midrule
        EDSR-baseline~\cite{lim2017enhanced}&32.00&26.04&0.784&32.00&27.71&0.782&32.00&28.80&0.814\\
        EDSR-baseline-PAMS~\cite{Li2020pams}&8.00&25.94&0.781&8.00&27.67&0.781&8.00&28.77&0.813\\
        EDSR-baseline-DAQ~\cite{hong2022daq}&4.00&25.73&0.772&4.00&27.60&0.777&4.00&28.67&0.809\\
        \rowcolor{orange!25}
        EDSR-baseline-CADyQ (Ours)&6.09&25.94&0.782& 5.52&27.67&0.781& 5.37&28.77&0.813\\
        \midrule
        SRResNet~\cite{ledig2017photo}&32.00&25.74&0.773&32.00&27.60&0.778&32.00&28.68&0.810\\
        SRResNet-PAMS~\cite{Li2020pams}&8.00&25.85&0.776&8.00&27.63&0.779&8.00&28.72&0.812\\
        SRResNet-DAQ~\cite{hong2022daq}&4.00&25.70&0.772&4.00&27.59&0.778&4.00&28.67&0.810\\
        \rowcolor{orange!25}
        SRResNet-CADyQ (Ours)&5.73&25.92&0.781& 5.14&27.64&0.781& 5.02&28.72&0.812\\
        \midrule
        CARN~\cite{ahn2018fast}&32.00&26.07&0.784&32.00&27.69&0.782&32.00&28.79&0.814\\
        CARN-PAMS~\cite{Li2020pams}&8.00&25.80&0.776&8.00&27.60&0.778&8.00&28.68&0.811\\
        CARN-DAQ~\cite{hong2022daq}&4.00&25.48&0.764&4.00&27.30&0.771&4.00&28.24&0.802\\
        \rowcolor{orange!25} 
        CARN-CADyQ (Ours)&5.32&25.94&0.780&4.65&27.65&0.780&4.54&28.73&0.812 \\
        \bottomrule
    \end{tabular}
}
\end{table}

\noindent\textbf{Models.}
The proposed framework is applied directly to existing SR networks, including representative SR networks (EDSR-baseline~\cite{lim2017enhanced} and SRResNet~\cite{ledig2017photo}) and recent efficient models (IDN~\cite{hui2018idn} and CARN~\cite{ahn2018fast}), thereby naming the CADyQ-quantized models as EDSR-baseline-CADyQ, SRResNet-CADyQ, IDN-CADyQ, and CARN-CADyQ, respectively.
Following the settings from previous works on SR quantization~\cite{Li2020pams,ma2019efficient,xin2020binarized}, our framework quantizes weights and feature maps in the layers of the high-level feature extraction module where most of the costly operations are concentrated in.
In this work, we set the quantization bit-width candidates as $\{4,6,8\}$ and employ linear symmetric quantization function~\cite{Li2020pams} with a learnable scale parameter for each quantization function candidate.
Furthermore, we uniformly apply 8-bit linear quantization for weights. 
Additional experiments that demonstrate the applicability of CADyQ are provided in supplementary document Sec.~\textcolor{red}{A}.
\\

\noindent\textbf{Training Details.}
Training and validation are done with DIV2K~\cite{agustsson2017ntire} dataset. 
For training stability, we follow~\cite{jung2019learning,zhuang2018towards} in initializing the SR network parameters with pre-trained $8$-bit network weights and in controlling the bit selector to progressively decrease the bit-width.
For a progressive reduction in bit-width, the weight of the proposed bit regularization loss ($w_{reg}$) is initially $10^{-4}$ and gradually increased throughout training ($10^{-6}$ per $1K$ iteration). 
The weights for the loss terms $w_1, w_\text{kd}$, and $w_\text{kdf}$ is respectively $1.0$, $1000.0$, and $100.0$.
Analysis on $w_{reg}$ and other training settings are specified in Sec.~\textcolor{red}{C} of the supplementary document. \\

\noindent\textbf{Evaluation Details.}
We evaluate our framework on the standard benchmark (Urban100~\cite{huang2015single}) and on more computationally demanding images of large size (e.g., 2K, 4K) in Test2K and Test4K datasets~\cite{kong2021classsr} which are generated via bicubic downsampling from DIV8K~\cite{gu2019div8k} dataset (index 1201-1400).
For testing, an input test image is cropped into patches of size $96\times96$ with six overlapping boundary pixels.
Each patch is super-resolved with our framework and then combined to produce the whole HR image.
There exists a trade-off between the patch size and the overall efficiency, which is discussed in Sec.~\ref{subsec:complexity} and supplementary document Sec.~\textcolor{red}{B}.
We report peak signal-to-noise ratio (PSNR) and structural similarity index (SSIM~\cite{wang2004image}) to evaluate the SR performance, along with the average feature quantization rate (FQR) for the evaluation of efficiency. 
Furthermore, our ablation study is conducted consistently with CARN backbone models on Urban100 dataset.

\subsection{Quantitative Results} \label{subsec:comparison}
To evaluate the effectiveness and efficiency of the proposed mechanism, 
we compare the results with PAMS~\cite{Li2020pams} and DAQ~\cite{hong2022daq} using the official code, which is similar to CADyQ in that quantization is directly applied to existing SR networks without redesigning the architecture.
Specifically, we compare with PAMS (8-bit) since lower-bit quantization by PAMS results in performance degradation, and DAQ (w4a4qq4) with 4-bit for weight and feature quantization.
As shown in Table~\ref{tab:exp-main}, DAQ reduces the computational resources but at the cost of performance degradation, which is especially severe (over -0.5dB) for IDN and CARN baseline.
Also, compared with PAMS, CADyQ demonstrates a lower average precision without the performance drop, striking a better balance between computational cost and performance.

\subsection{Qualitative Results} \label{subsec:comparison-qual}
\begin{figure}[t]
\begin{center}
    \centering
    \setlength{\tabcolsep}{0.07cm}
    \newcommand{\w}{0.185\linewidth}
    \begin{tabular}{ccccc}
        \includegraphics[width=\w]{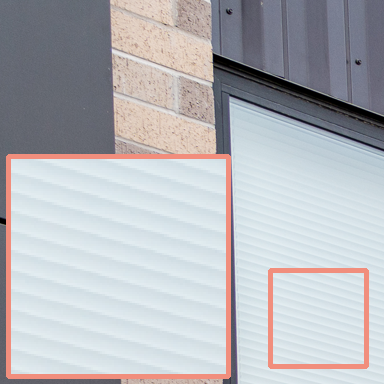} &
        \includegraphics[width=\w]{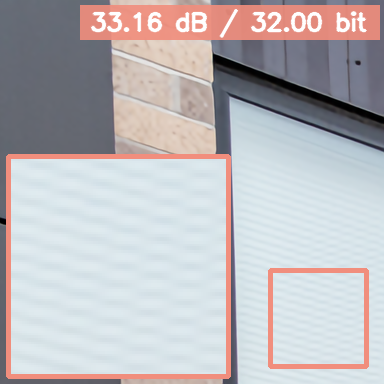} &
        \includegraphics[width=\w]{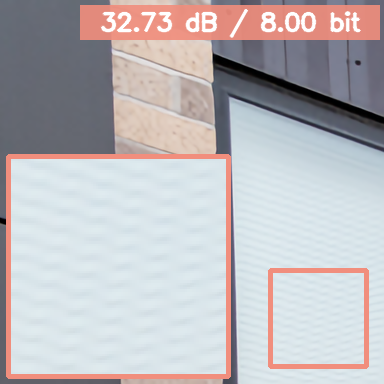} &
        \includegraphics[width=\w]{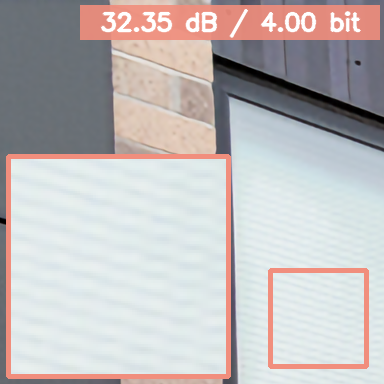} &
        \includegraphics[width=\w]{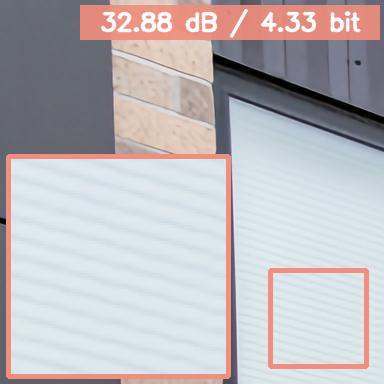} \\
        \scriptsize GT &\scriptsize CARN~\cite{ahn2018fast} &\scriptsize CARN-PAMS~\cite{Li2020pams} &\scriptsize CARN-DAQ~\cite{hong2022daq} &\scriptsize CARN-CADyQ \\
        \includegraphics[width=\w]{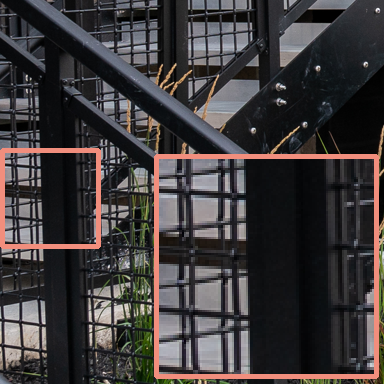} &
        \includegraphics[width=\w]{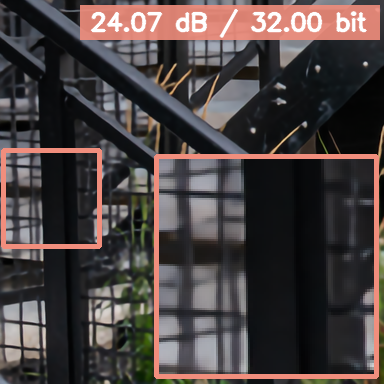} &
        \includegraphics[width=\w]{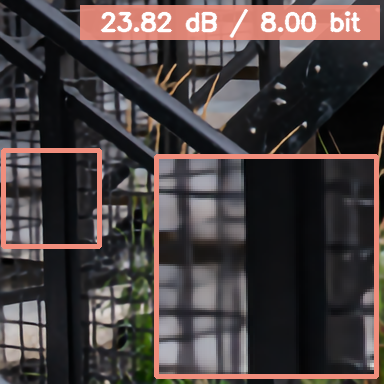} &
        \includegraphics[width=\w]{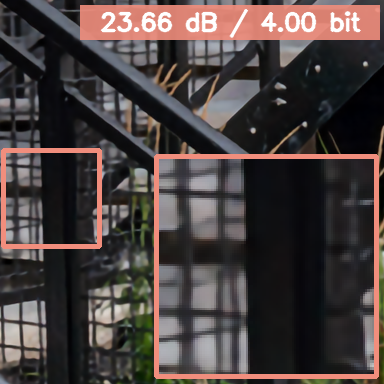} &
        \includegraphics[width=\w]{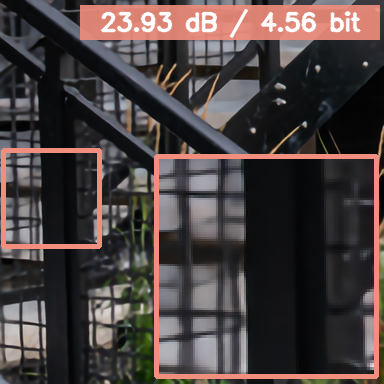} \\
        \scriptsize GT & \scriptsize CARN~\cite{ahn2018fast} &\scriptsize CARN-PAMS~\cite{Li2020pams} &\scriptsize CARN-DAQ~\cite{hong2022daq}& \scriptsize CARN-CADyQ 
        \\
    \end{tabular}
    \caption{
    \textbf{Qualitative results} on `img1215' of Test4K. Quantitative measures of PSNR and average bit-width of the patch are also reported (PSNR / Average bit-width).
    More results are provided in Section~\textcolor{red}{E} of the supplementary document
    }
    \label{fig:exp-qualitative}
\end{center}
\end{figure}

\begin{figure}[t]
    \includegraphics[width=1.0\textwidth]{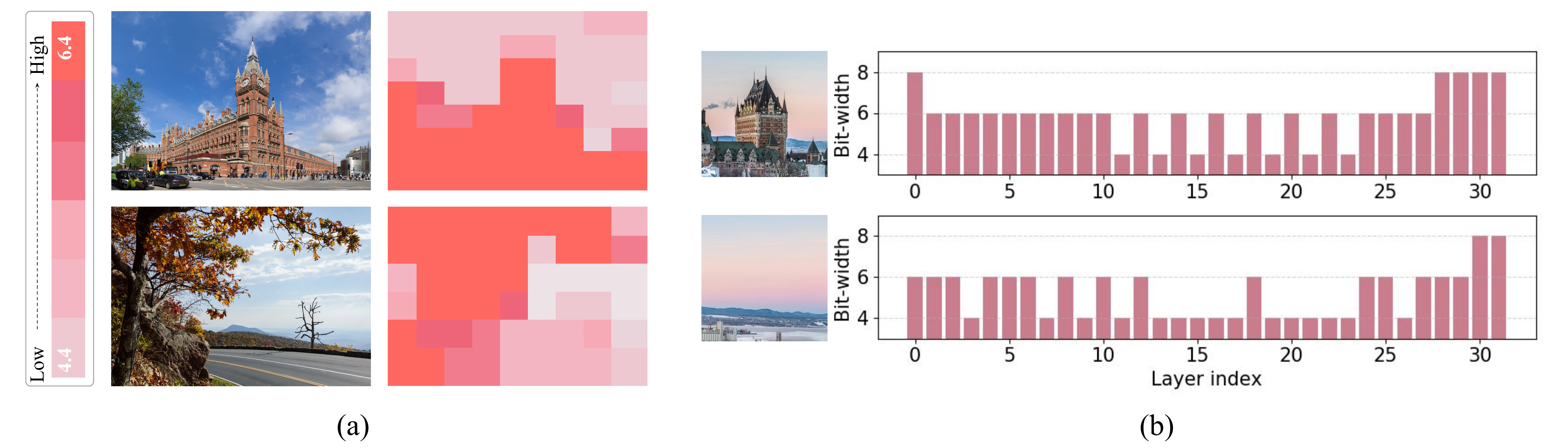}
    \caption{
        \textbf{Visualizations of dynamic bit-width allocation by CADyQ across patches and layers.} 
        On average, CADyQ assigns higher bit-width to \textbf{(a)} complex patches and to \textbf{(b)} important layers.
        Results are obtained with EDSR-CADyQ from (a) `img1215' and `img1222' from Test4K and (b) two patches of `img1400' from Test2K
    }
    \label{fig:exp-visualization}
\end{figure}

Fig.~\ref{fig:exp-qualitative} provides qualitative results and comparisons with the output images from CARN-based models~\cite{ahn2018fast}.
CARN-CADyQ (ours) produces a visually clean output image, while CARN-PAMS and sometimes even original unquantized CARN suffer from a checkerboard artifact or blurred lines, even though CARN-CADyQ uses less computational resources. 
Moreover, CARN-DAQ, despite the low computational resources, produces various artifacts and color distortion.
Also, the map of average bit-width used by our framework for each local patch in the image is visualized in Fig.~\ref{fig:exp-visualization} (a).
The visualized bit map demonstrates that our framework allocates more computational resources to patches with complex structures (e.g., buildings) and less to patches with simple structures (e.g., sky).
Furthermore, CADyQ is shown to dynamically allocate distinct bit-widths across different network layers, as visualized in Fig.~\ref{fig:exp-visualization} (b).
The qualitative results stress the effectiveness and importance of the patch-and-layer-wise bit allocation.

\subsection{Ablation Study}\label{subsec:ablation}

\definecolor{brickred}{rgb}{0.8, 0.25, 0.33}
\definecolor{brickred2}{rgb}{0.25, 0.8, 0.33}
\newcommand{\cm}{\color{brickred2}{\ding{51}}}%
\newcommand{\xm}{\color{brickred}{\ding{55}}}%

\begin{table}[t]
    \centering
    \caption{ \textbf{Ablation study on layer-wise and patch-wise quantization}
    }
    \label{tab:exp-ablation-1}
    \begin{tabular}{@{}c|ccccc }
        \toprule
        & Layer-wise & Patch-wise & FQR$_\downarrow$ & PSNR$_\uparrow$ &SSIM$_\uparrow$\\
        \midrule
        \textbf{(2a)} & \xm & \xm & 8.00&25.80 &0.776\\
        \textbf{(2b)} & \xm & \cm & 6.15&25.89 &0.778\\
        \textbf{(2c)} & \cm & \xm & 7.02&25.92 &0.780\\\rowcolor{orange!25}
        \textbf{CADyQ} & \cm & \cm & 5.32&25.94 &0.780\\
        \bottomrule
    \end{tabular}

\end{table}

\noindent\textbf{Effect of Layer-Wise and Patch-Wise Quantization.}
To verify the importance of layer-wise and patch-wise quantization in conjunction, we compare our overall scheme CADyQ with its separate modules individually: patch-wise quantization and layer-wise quantization, as reported in Table~\ref{tab:exp-ablation-1}.
Quantization with patch-wise dynamic bit-width but fixed throughout the network (model \textbf{(2b)}) results in a performance drop.
A layer-wise different bit but fixed across different patches and images (model \textbf{(2c)}) preserves the restoration accuracy but with a small improvement in efficiency (average 7.02 bit).
By contrast, layer-wise and patch-wise quantization in conjunction effectively enhances the quality of the super-resolved image and reduces the average bit-width by a large amount.
The results validate our claim that dynamically determining the bit-width both per layer and patch is important, corroborating our observations from Fig.~\ref{fig:method-observation}.  \\
\begin{table}[t]
    \centering
    \caption{ \textbf{Ablation study on quantization sensitivity measures}
    }
    \scalebox{0.9}{
        \begin{tabular}{@{}c|ccccc}
            \toprule
            & Patch & Layer & $\text{FQR}_\downarrow$ & $\text{PSNR}_\uparrow$ & $\text{SSIM}_\uparrow$\\
            \midrule
            \textbf{(3a)} & max-min & max-min & 4.51 & 25.16&0.752\\
            \textbf{(3b)} & std & layer std & 5.95 & 25.62&0.769 \\
            \textbf{(3c)} & gradient & layer std & 6.53 & 25.80&0.775\\\rowcolor{orange!25}
            \textbf{CADyQ} & gradient & channel std & 5.32 & 25.94&0.780\\
            \bottomrule
        \end{tabular}
    }
    \label{tab:exp-ablation-2}
\end{table}

\noindent\textbf{Quantization Sensitivity Estimation.}
In this ablation study, we validate our choice of measures for quantization sensitivity, which a bit selector uses to decide the bit-width for each patch and layer, as shown in Table~\ref{tab:exp-ablation-2}.
We compare alternative measures that could estimate the quantization sensitivity and have similar computational overheads to our choice: patch gradient and channel-wise standard deviation.
Although utilizing the range (the gap between max and min value) of the patch pixel values and the layer-wise feature (model \textbf{(3a)}) requires fewer computations, it induces a severe performance degradation.
Also, the standard deviation of a patch and the layer-wise standard deviation of the feature (model \textbf{(3b)}) suffers from performance degradation.
Notably, using the layer-wise standard deviation (model \textbf{(3c)}) results in lower performance (PSNR or SSIM) and higher average precision (FQR), compared with using channel-wise standard deviation (CADyQ).
We provide further justifications for using standard deviations of each channel in the supplementary document Sec.~\textcolor{red}{B}.
In summary, the ablation study implies that the patch gradient and channel-wise standard deviation of the feature contain important information that gives a better estimate of the quantization sensitivity,
thereby helping to find a better trade-off between performance and computational resources. \\

\begin{table}[t]
    \begin{minipage}{.49\linewidth}
        \centering
        \caption{\textbf{Ablation study on  losses:} bit loss and knowledge distillation loss
        }
        \label{tab:exp-ablation-3}
        \scalebox{0.65}{
        \small
        \begin{tabular}[b]{@{}c|cccc ccc}
        \toprule
        & \multicolumn{4}{c}{Loss}& \multirow{2}{*}{FQR$_\downarrow$} & \multirow{2}{*}{PSNR$_\uparrow$} & \multirow{2}{*}{SSIM$_\uparrow$}\\
        \cmidrule{2-5} 
        & $\mathcal{L}_1$ & $\mathcal{L}_\text{reg}$ & $\mathcal{L}_\text{kd}$ & $\mathcal{L}_\text{kdf}$ \\
        \midrule
        \textbf{(4a)} & \cm &          \xm            & \cm & \cm & 6.51 & 25.70 & 0.772 \\
        \textbf{(4b)} & \cm & $\mathcal{L}_\text{b}$  & \cm & \cm & 5.75 & 25.68 & 0.772 \\
        \textbf{(4c)} & \cm & $\mathcal{L}_\text{wb}$ & \xm & \cm & 4.48 & 25.38 & 0.761 \\
        \textbf{(4d)} & \cm & $\mathcal{L}_\text{wb}$ & \cm & \xm & 5.25 & 25.51 & 0.766 \\
        \rowcolor{orange!25}
        \textbf{CADyQ} & \cm & $\mathcal{L}_\text{wb}$ & \cm & \cm & 5.32 & 25.94 & 0.780 \\
        \bottomrule
    \end{tabular}
    }
    \end{minipage} 
    \hfill
    \begin{minipage}{.49\linewidth}
        \centering
        \includegraphics[width=0.9\linewidth]{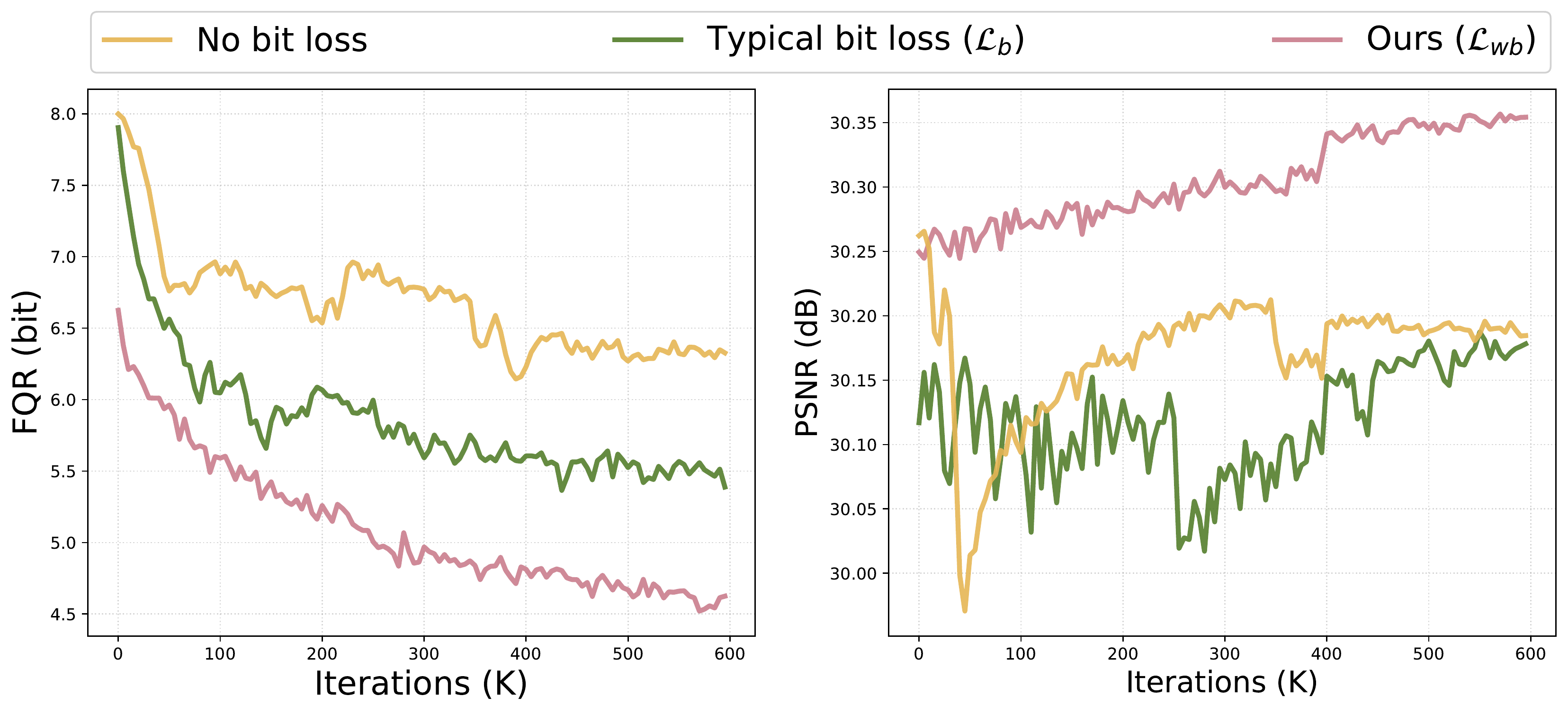}
        \captionof{figure}{\textbf{Learning curves with different losses} for FQR (left) and PSNR (right)
        }
        \label{fig:exp-loss}
    \end{minipage} 
\end{table}

\noindent\textbf{Ablation on Losses.}
We analyze the effect of the proposed weighted bit loss $\mathcal{L}_\text{wb}$ (Eq.~\eqref{eqn:method-bitloss5}) by removing it (model \textbf{(4a)}) or replacing it with the conventional bit loss $\mathcal{L}_\text{b}$, formulated in Eq.~\eqref{eqn:method-bitloss-c}, (model \textbf{(4b)}) in the overall objective function from Eq.~\eqref{eqn:method-loss}, as displayed in Table~\ref{tab:exp-ablation-3} and  Fig.~\ref{fig:exp-loss}.
The figure shows the curves of the average bit-widths and PSNR of the validation dataset during training.
Without the bit loss, the framework fails to reduce the computational resources effectively.
Also, replacing our bit loss with the conventional bit loss reduces the average bit-width but with $\sim0.2$ dB PSNR drop.
On the other hand, our proposed bit loss substantially reduces the computational resources without a PSNR drop.
Then we evaluate the effect of knowledge distillation loss by removing the distillation loss on the output feature (model \textbf{(4c)}) and intermediate features (model \textbf{(4d)}), respectively.
The results in the table suggest that both types of knowledge distillation play key roles in maintaining the restoration performance. 

\begin{table}[t]
    \begin{minipage}{.49\linewidth}
        \centering
        \caption{\textbf{Complexity analysis of image-wise and patch-wise inference} 
        measured w.r.t. BitOPs of the feature extraction stage on CARN backbone models
        }
        \label{tab:exp-complexity}
        \scalebox{0.65}{
        \small
        \begin{tabular}[b]{@{}c|ccc|ccc}
            \toprule
            Patch Size & \multicolumn{3}{c}{Full Image} |&  \multicolumn{3}{c}{96$\times$96} \\ 
            \midrule
            Model & Baseline & PAMS & \textbf{CADyQ} & Baseline & PAMS & \textbf{CADyQ}\\
            \midrule
            PSNR$_\uparrow$(dB) & 26.07 & 25.80 & \textbf{25.95} & 26.07 & 25.80 & \textbf{25.94}\\
            BitOPs$_\downarrow$(G) &76.44 & 4.78 & \textbf{3.24} & 77.87 & 4.87 & \textbf{3.23}\\
            \bottomrule
        \end{tabular}
    }
    \end{minipage} 
    \hfill
    \begin{minipage}{.49\linewidth}
        \centering
        \caption{
        \textbf{Average GPU inference latency.} GPU latency for each model is measured with EDSR backbone models on Test4K images
        }
        \label{tab:exp-latency}
        \scalebox{0.9}{
        \small
        \begin{tabular}[b]{@{}c|ccc}
            \toprule
            Model & Baseline & PAMS & \textbf{CADyQ}\\
            \midrule
            GPU Inference & \multirow{2}{*}{535.5} & \multirow{2}{*}{240.0}& \multirow{2}{*}{\textbf{206.5}}\\
            Latency (ms) & &&\\
            \bottomrule
        \end{tabular}
        }
    \end{minipage} 
\end{table}

\subsection{Complexity Analysis}
\label{subsec:complexity}
Our framework can process either the full input test image at once or process, in parallel, the smaller patches, which are combined to construct the full image.
For both scenarios, we analyze the complexity of our framework w.r.t. the number of operations weighted by the bit-widths of the operands (BitOPs)
for generating a $720$p ($1820\times720$) image in Table~\ref{tab:exp-complexity}.
We use BitOPs as measurements for the computational complexity to better reflect the reduction in bit-width.
Our framework is shown to be more effective on the patch-wise inference, as local regions with complex structures and those with simple structures are processed with different computational resources.
Despite additional computational overhead from overlapping area between neighboring patches in the patch-wise inference, our framework achieves $\sim$95.8\% reduction in BitOPs compared with the baseline and $\sim$32.2\% reduction in BitOPs compared with 8-bit CARN-PAMS at image-wise inference.
GPU latency is measured on NVIDIA Tesla T4 GPU with Tensor Cores supporting 4/8-bit acceleration.
As hardware-acceleration for 6-bit is not supported on T4, we cast 6-bit assignments as 8-bit.
On average, inference latency of Test4K images is improved to 206.5ms for CADyQ, compared with 240.0ms for 8-bit quantization (PAMS~\cite{Li2020pams}), and 535.5ms for 32-bit.
Computational complexity of other backbone models and detailed analysis of overheads can be found in Sec.~\textcolor{red}{D} of the supplementary document.
\section{Conclusion}
\label{sec:conclusion}
In this work, we study and exploit the relationship between the local image contents (\textit{e.g.}, local patches and their features at each layer) and the super-resolution performance degradation from quantization. 
We thereby propose a patch-and-layer-wise bit allocation method for dynamic quantization.
Experimental results demonstrate that the proposed quantization framework, CADyQ manages to reduce the computational complexity with respect to BitOPs and inference latency with negligible performance drop.
\section*{Acknowledgment}
\label{sec:acknowledgment}
This work was supported in part by the IITP grant funded by the Korea government (MSIT) [No. 2021-0-01343, Artificial Intelligence Graduate School Program (Seoul National University), No. 2021-0-02068, Artificial Intelligence Innovation Hub, and No.2022-0-00156], and in part by the BK21 FOUR program of the Education and Research Program for Future ICT Pioneers, Seoul National University in 2022.

\clearpage
\bibliographystyle{splncs04}
\bibliography{egbib}
\clearpage

\end{document}


\pagestyle{headings}
\mainmatter
\def\ECCVSubNumber{5937}

\title{
CADyQ: Content-Aware Dynamic Quantization for Image Super-Resolution \\
-- \textit{Supplementary Document} -- 
}

\titlerunning{CADyQ: Content-Aware Dynamic Quantization for Image Super-Resolution}

\author{Cheeun Hong\inst{1} \and
Sungyong Baik\inst{3} \and
Heewon Kim\inst{1} \and \\
Seungjun Nah\inst{1,4} \and
Kyoung Mu Lee\inst{1,2}
}

\authorrunning{C. Hong et al.}

\institute{
Dept. of ECE \& ASRI, \email{\{cheeun914, ghimhw, kyoungmu\}@snu.ac.kr} \and
IPAI, Seoul National University \and
Dept. of Data Science, Hanyang University \email{dsybaik@hanyang.ac.kr} \and
NVIDIA \email{seungjun.nah@gmail.com} 
}

\maketitle
This supplementary document presents \textbf{additional experimental quantitative results} to further demonstrate the effectiveness of our framework in Section~\ref{sec:sup-exp}; \textbf{additional analyses} of our proposed framework in Section~\ref{sec:sup-analyses}; \textbf{implementation details} in Section~\ref{sec:sup-implementation}; \textbf{complexity analysis} in Section~\ref{sec:sup-complexity}; and \textbf{additional qualitative results} in Section~\ref{sec:sup-qualitative}.

\section{Additional Experiments} \label{sec:sup-exp}

\subsection{Effectiveness on Other Quantization Function}
In this section, we provide experimental results to demonstrate that the proposed quantization framework is applicable to other quantization functions.
In the main text, each quantization function candidate is implemented with PAMS, a linear quantization function with a learnable scale parameter~\citenumber{31}.
To demonstrate the generalizability of CADyQ with different quantization functions in this section, we employ another linear quantization function~\cite{krishnamoorthi2018whitepaper}, which we refer to as LinQ.
The major difference between PAMS and LinQ is that PAMS uses a learnable scale parameter to clip the outliers, whereas LinQ does not clip the outliers and thus takes the whole feature range as the quantization range.
Table~\ref{tab:sup-linq} shows the quantization results of CADyQ with LinQ (referred to as {LinQ-CADyQ}).
Our framework successfully achieves the reduction in average precision while simultaneously achieving the performance similar to or better than baselines or its 8-bit counterpart quantized with LinQ.
The results show that the proposed quantization framework is flexible in terms of quantization functions.
Note that CADyQ achieves a larger average precision reduction with PAMS than with LinQ.
The results corroborate the previous findings that learnable scale parameters are effective for SR networks, which can have variant outliers in feature distributions~\citenumber{31}.
\begin{table}[h!]
\caption{\textbf{Quantitative comparisons with other quantization function.}
The results demonstrate the effectiveness of the proposed framework on SR networks (IDN~\citenumber{21}, EDSR-baseline~\citenumber{34}, SRResNet~\citenumber{29}, and CARN~\citenumber{2}) regardless of the quantization function type
}
\centering
\resizebox{1.\linewidth}{!}{
    \begin{tabular}{l rcc rcc rcc}
        \toprule
        \multirow{2}{*}{Model} & 
        \multicolumn{3}{c}{Urban100} & \multicolumn{3}{c}{Test2K} & \multicolumn{3}{c}{Test4K} \\
        \cmidrule(lr){2-4} \cmidrule(lr){5-7} \cmidrule(lr){8-10} & 
        FQR$_\downarrow$ & PSNR$_\uparrow$ & SSIM$_\uparrow$ & 
        FQR$_\downarrow$ & PSNR$_\uparrow$ & SSIM$_\uparrow$ & 
        FQR$_\downarrow$ & PSNR$_\uparrow$ & SSIM$_\uparrow$\\
        \midrule
        IDN~\citenumber{21}               &32.00 &25.42 &0.763 &32.00 &27.48 &0.774 &32.00 &28.54 &0.806\\
        IDN-LinQ~\cite{krishnamoorthi2018whitepaper}& 8.00 &25.47 &0.764 & 8.00 &27.50 &0.774 & 8.00 &28.56 &0.806\\\rowcolor{orange!25}
        IDN-{LinQ-CADyQ} (Ours)      & 6.85 &25.60 &0.769 & 6.65 &25.72 &0.775 & 6.57 &28.58 &0.807\\
        \midrule
        EDSR-baseline~\citenumber{34}                  &32.00 &26.04 &0.784 &32.00 &27.71 &0.782 &32.00 &28.80 &0.814\\
        EDSR-baseline-LinQ~\cite{krishnamoorthi2018whitepaper}   & 8.00 &25.85 &0.777 & 8.00 &27.65 &0.780 & 8.00 &28.74 &0.812\\\rowcolor{orange!25}
        EDSR-baseline-LinQ-CADyQ (Ours)                          & 7.26 &25.96 &0.782 & 7.20 &27.68 &0.781 & 7.20 &28.78 &0.813\\
        \midrule
        SRResNet~\citenumber{29}                  &32.00 &25.74 &0.773 &32.00 &27.60 &0.778 &32.00 &28.68 &0.810\\
        SRResNet-LinQ~\cite{krishnamoorthi2018whitepaper}   & 8.00 &25.83 &0.777 & 8.00 &27.62 &0.780 & 8.00 &28.70 &0.812\\\rowcolor{orange!25}
        SRResNet-LinQ-CADyQ (Ours)                          & 6.49 &25.87 &0.777 & 6.38 &27.61 &0.780 & 6.33 &28.69 &0.811\\
        \midrule
        CARN~\citenumber{2}                   &32.00 &26.07 &0.784 &32.00 &27.69 &0.782 &32.00 &28.79 &0.814\\
        CARN-LinQ~\cite{krishnamoorthi2018whitepaper}   & 8.00 &25.82 &0.776 & 8.00 &27.60 &0.779 & 8.00 &28.68 &0.811\\\rowcolor{orange!25}
        CARN-LinQ-CADyQ (Ours)                          & 6.24 &25.88 &0.778 & 6.00 &27.63 &0.779 & 5.83 &28.71 &0.811\\
        \bottomrule
    \end{tabular}
}
\label{tab:sup-linq}
\end{table}

\subsection{Experiments on SR Networks for Scale $\times$ 2}
In addition to the scale $\times 4$ experiments presented in the main manuscript, we additionally evaluate our framework on SR models for scale $\times 2$ : IDN~\citenumber{21}, EDSR-baseline~\citenumber{34}, SRResNet~\citenumber{29}, and CARN~\citenumber{2}.
As shown in Table~\ref{tab:sup-scale2}, compared to DAQ~\citenumber{17} (4-bit) that suffers from severe performance degradation, CADyQ achieves minimal or no performance degradation from the unquantized baseline model.
Compared to PAMS~\citenumber{31} (8-bit), the average feature quantization rate (FQR) is reduced while achieving a similar or better performance, demonstrating the effectiveness of our framework on scale $\times 2$.
\begin{table}[h!]
\centering
\caption{ \textbf{Quantitative comparisons on various SR networks:}
IDN~\citenumber{21}, EDSR-baseline~\citenumber{34}, SRResNet~\citenumber{29}, and CARN~\citenumber{2} of \textbf{scale $\times$ 2}
}
\resizebox{1.\linewidth}{!}{
    \begin{tabular}{l rcc rcc rcc}
        \toprule
        \multirow{2}{*}{Model} & 
        \multicolumn{3}{c}{Urban100} & \multicolumn{3}{c}{Test2K} & \multicolumn{3}{c}{Test4K} \\
        \cmidrule(lr){2-4} \cmidrule(lr){5-7} \cmidrule(lr){8-10} & 
        FQR$_\downarrow$ & PSNR$_\uparrow$ & SSIM$_\uparrow$ & 
        FQR$_\downarrow$ & PSNR$_\uparrow$ & SSIM$_\uparrow$ & 
        FQR$_\downarrow$ & PSNR$_\uparrow$ & SSIM$_\uparrow$\\
        \midrule
        IDN~\citenumber{21}       &32.00 &31.29 &0.920 &32.00 &32.42 &0.924 &32.00 &34.02 &0.940\\
        IDN-PAMS~\citenumber{31}  &8.00 &31.39 &0.921 &8.00 &32.46 &0.925 & 8.00 &34.05 &0.941\\
        IDN-DAQ~\citenumber{17}   &4.00 &29.15 &0.886 &4.00 &31.33 &0.904 & 4.00 &32.70 &0.921\\
        \rowcolor{orange!25}
        IDN-CADyQ (Ours)                    &5.22 &31.54 &0.923 &4.67 &32.51 &0.925 &4.57 &34.10 &0.941\\
        \midrule
        EDSR-baseline~\citenumber{34}      &32.00 &31.97 &0.927 &32.00 &32.75 &0.928 &32.00 &34.37 &0.943\\
        EDSR-baseline-PAMS~\citenumber{31} &8.00 &31.96 &0.927 &8.00 &32.72 &0.928 &8.00 &34.33 &0.943\\
        EDSR-baseline-DAQ~\citenumber{17}  &4.00 &31.63 &0.923 &4.00 &32.58 &0.926 &4.00 &34.15 &0.942\\
        \rowcolor{orange!25}
        EDSR-baseline-CADyQ (Ours)                   &6.15 &31.95 &0.927 &5.68 &32.70 &0.928 &5.59 &34.30 &0.943\\
        \midrule
        SRResNet~\citenumber{29}     &32.00 &31.44 &0.921 &32.00 &32.51 &0.925 &32.00 &34.11 &0.941\\
        SRResNet-PAMS~\citenumber{31} &8.00 &31.44 &0.922 &8.00 &32.52 &0.925 &8.00 &34.11 &0.941\\
        SRResNet-DAQ~\citenumber{17}  &4.00 &31.25 &0.919 &4.00 &32.42 &0.924 &4.00 &33.98 &0.940\\
        \rowcolor{orange!25}
        SRResNet-CADyQ (Ours)                  &6.46 &31.58 &0.923 &6.10 &32.61 &0.926 &6.02 &34.19 &0.942\\
        \midrule
        CARN~\citenumber{2}       &32.00 &31.92 &0.926 &32.00 &32.75 &0.928 &32.00 &34.32 &0.943\\
        CARN-PAMS~\citenumber{31} &8.00 &31.74 &0.924 &8.00 &32.62 &0.927 & 8.00 &34.22 &0.942\\
        CARN-DAQ~\citenumber{17}  &4.00 &30.74 &0.912 &4.00 &31.89 &0.918 & 8.00 &33.51 &0.936\\
        \rowcolor{orange!25}
        CARN-CADyQ (Ours)                   &4.32 &31.87 &0.925 &4.04 &32.66 &0.927 &4.03 &34.25 &0.942\\
        \bottomrule
    \end{tabular}
}
\label{tab:sup-scale2}
\end{table}

\clearpage
\subsection{Fully Quantized Super-Resolution Networks}
In most SR quantization works~\citenumber{31, 46}, not all layers were quantized.
Specifically, the shortcut connections in ResBlocks, the first and the last convolutional layers, were not quantized as they are known to have high quantization sensitivity.
FQSR~\citenumber{41} is the first work to quantize all layers and the shortcut connections in SR networks.
In Table~\ref{tab:sup-fq}, we show that CADyQ substantially reduces the average precision while maintaining the performance even when quantizing with FQSR on layers, including the ones with high quantization sensitivity.
The results outline the effectiveness of the proposed quantization sensitivity estimation measures (\textit{i.e.}, the average gradient magnitude of the given patch and the standard deviation of the layer feature) and the dynamic bit-width selection in the CADyQ framework.
\begin{table}[h!]
\centering
\caption{
\textbf{Quantitative comparison with fully quantized SR network.}
Evaluation is conducted with EDSR-baseline~\citenumber{34} as a backbone, on Urban100.
The weights of the fully quantized networks are quantized to 8-bit. 
Computational cost is measured w.r.t FQR which denotes the average feature quantization rate of the quantized layers and BitOPs which is calculated over the whole network
}
\scalebox{0.9}{
    \begin{tabular}{l rrcc}
        \toprule
        {Model} & 
        FQR$_\downarrow$ & BitOPs$_\downarrow$ & PSNR$_\uparrow$ & SSIM$_\uparrow$ \\
        \midrule
        EDSR-baseline~\citenumber{34} &32.00 &317.5G &26.04 &0.784\\
        EDSR-baseline-FQSR~\citenumber{41} & 8.00 &19.8G &25.92 &0.781\\\rowcolor{orange!25}
        EDSR-baseline-FQSR-CADyQ                     & 5.90 &17.6G &25.90 &0.779\\
        \bottomrule
    \end{tabular}
}
\label{tab:sup-fq}
\end{table}

\section{Additional Analyses} \label{sec:sup-analyses}
\subsection{Analysis on Quantization Sensitivity Estimation}
\begin{figure*}[h!]      
    \centering
    \newcommand{\wwp}{0.49\linewidth}
    \centering
    \subfloat[Channel-wise feature of 16-th layer in EDSR \label{fig:sup-measure-a}]{\includegraphics[width=\wwp]{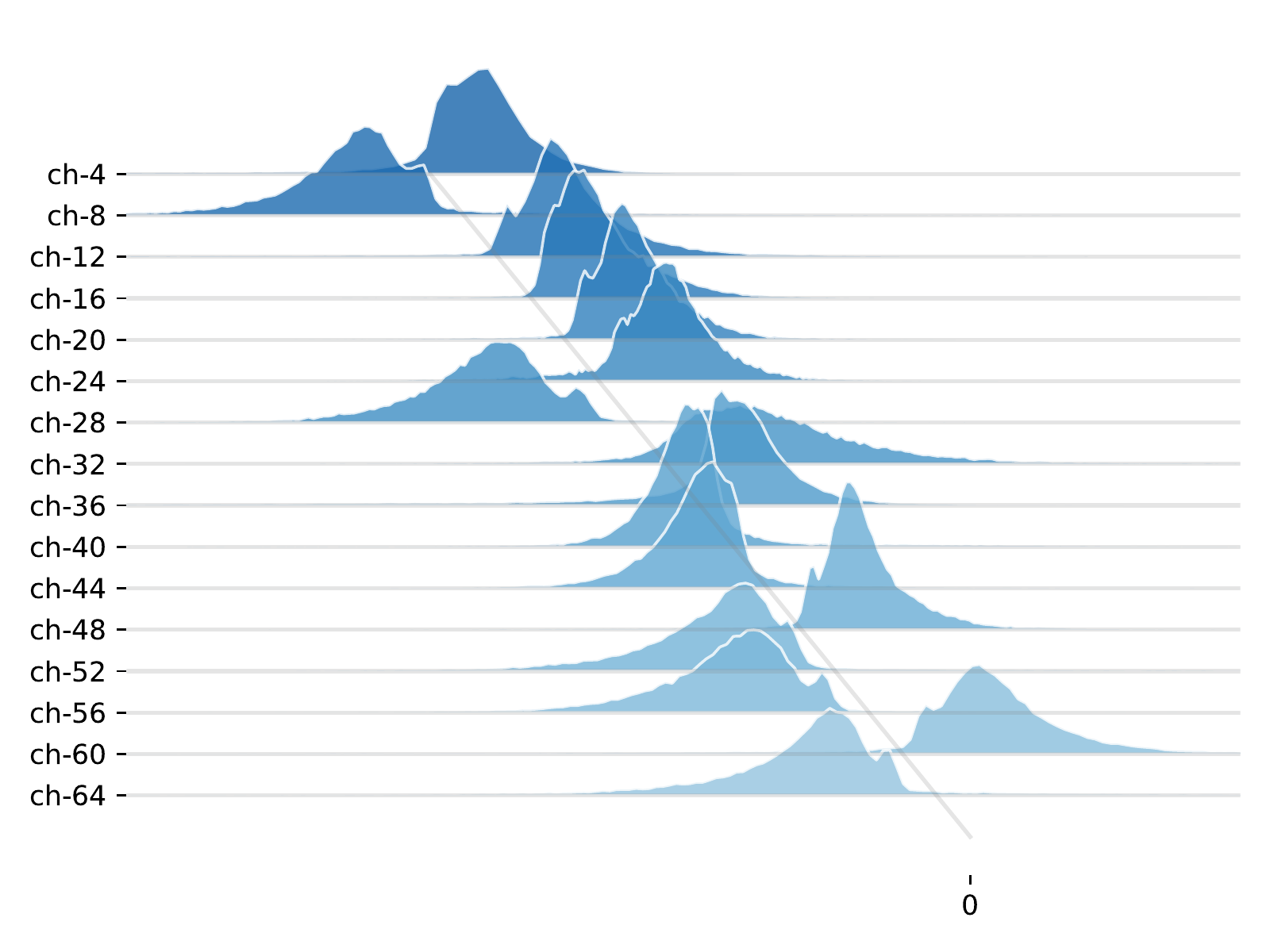}
    }
    \hfill
    \subfloat[Channel-wise vs layer-wise standard deviation of each layer \label{fig:sup-measure-b}]{\includegraphics[width=\wwp]{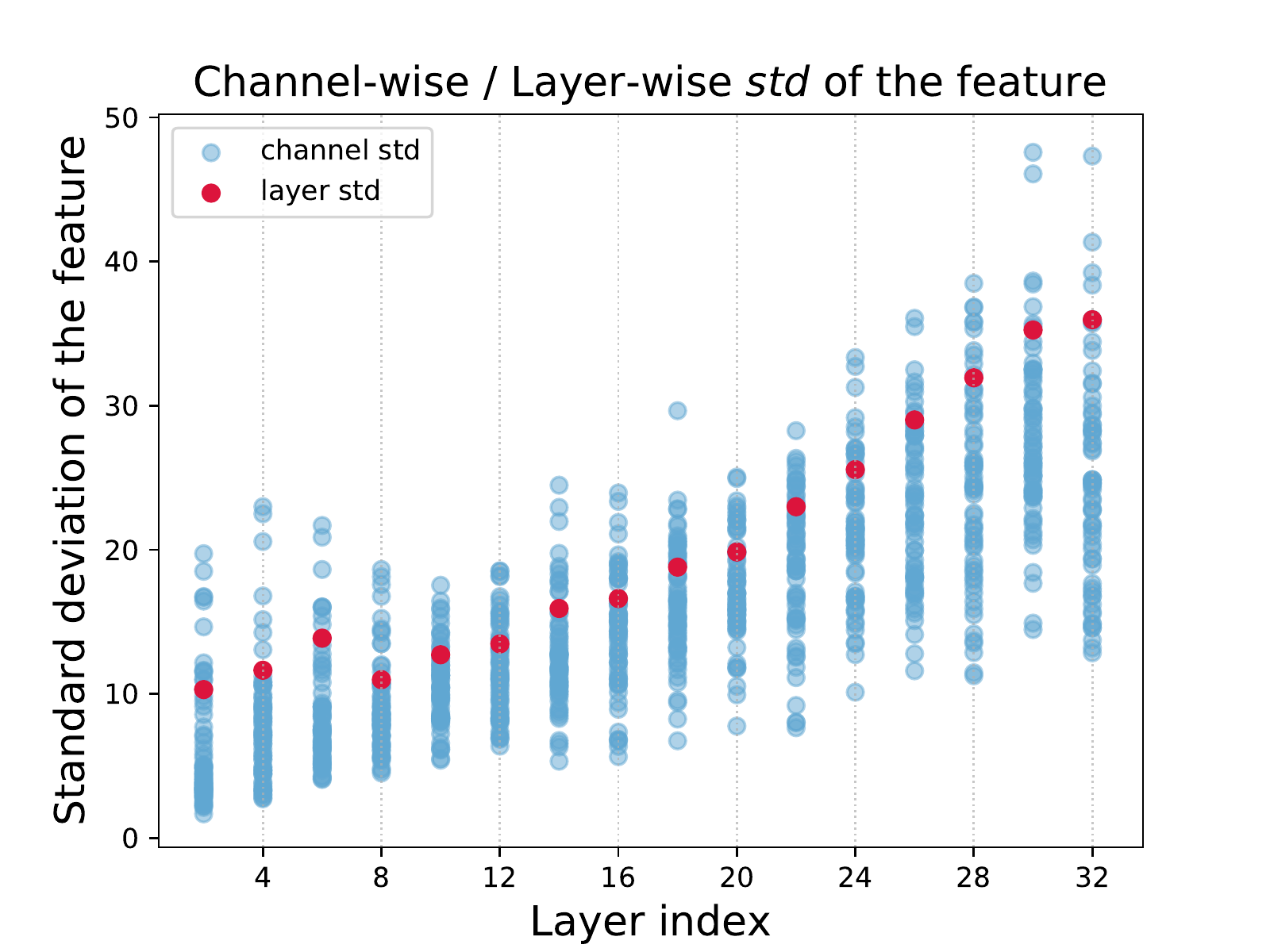}
    }
    \caption{
        \textbf{Visualization of feature distributions in SR networks}
    }
    \label{fig:sup-measure}
\end{figure*}

The results reported in Table~\textcolor{red}{3} of the main paper justify the use of the channel-wise standard deviation to estimate the quantization sensitivity of each layer feature.
Fig.~\ref{fig:sup-measure-a} shows that SR network layers have distinct channel distributions, where the $16$-th layer feature distributions of EDSR are illustrated as an example.
As each channel has a distinct distribution, the channels have diverse standard deviations, as demonstrated in Fig.~\ref{fig:sup-measure-b}.
Thus, using layer-wise standard deviation would result in loss of such information on distinct channel distributions, thereby obtaining less accurate quantization sensitivity estimation.

\subsection{The Variation of Quantization Bit-Width Candidates}
In this section, we investigate the impact of using different combinations of candidate bit-widths, as displayed in Table~\ref{tab:sup-searchspace}.
The results demonstrate that using different combinations of candidates results in different trade-offs between the computational complexity (FQR) and the restoration performance (PSNR and SSIM).
Such model behaviors can provide us the flexibility in selecting a model with the desired computational complexity and performance to target for specific applications, ranging from mobile devices to 4K displays.
For instance, adding a candidate bit-width of $6$~(CADyQ,~\textbf{(S5b)}) to the candidate bit-widths $\{4,8\}$~\textbf{(S4a)} leads to higher restoration performance but also higher computational complexity.
On the other hand, adding a candidate bit-width of $2$~\textbf{(S5c)} to the current candidate bit-widths $\{4,6,8\}$ results in lower computational complexity but also lower restoration performance.

\begin{table}[!h]
\centering
\caption{ \textbf{Ablation study on variation of quantization bit-width candidates.}
Evaluation is conducted by utilizing CARN as a backbone, on Urban100
}
\scalebox{0.9}{
    \begin{tabular}{@{}c|cccc }
        \toprule
        & Candidate bit-widths & FQR$_\downarrow$ & PSNR$_\uparrow$ &SSIM$_\uparrow$\\
        \midrule
        \textbf{(S5a)} & \{4, 8\}& 5.02 & 25.90 & 0.779 \\
        \textbf{(S5b)} & \{4, 6, 8\} & 5.32 & 25.94 & 0.780 \\
        \textbf{(S5c)} & \{2, 4, 6, 8\} & 3.45 & 25.61 & 0.770 \\
        \bottomrule
    \end{tabular}
}
\label{tab:sup-searchspace}
\end{table}

\subsection{Ablation on Different Inference Patch Size}
Larger patches have overall similar average magnitude of image gradient (\textit{i.e.}, smaller variance).
As CADyQ is conditioned on the average gradient magnitude of each patch, smaller variance prevents CADyQ from allocating distinct bit-widths.
Thus, BitOPs are less reduced for large patches, though still low compared to the existing methods, as in Table~\ref{tab:reb-patch}. 
By contrast, smaller patches further reduce BitOPs and maintain comparable PSNR.

\begin{table}[!h]
\centering
\caption{ \textbf{Ablation on patch size} on CARN-CADyQ, Urban100}
\scalebox{0.9}{
    \begin{tabular}{c |c|cc}
        \toprule
        Patch size&  Variance of $|\nabla{I_i}|$ &PSNR (dB) & BitOPs (G)  \\
        \midrule
        192$\times$192 & 1.41 & 25.95 & 3.36\\
        96$\times$96   & 1.79 & 25.94 & 3.23\\
        48$\times$48   & 2.53 & 25.92 & 3.12\\
        \bottomrule
    \end{tabular}
}
\label{tab:reb-patch}
\end{table}

\subsection{Details on $w_{reg}$}
\begin{figure}
    \centering
    \includegraphics[width=0.4\textwidth]{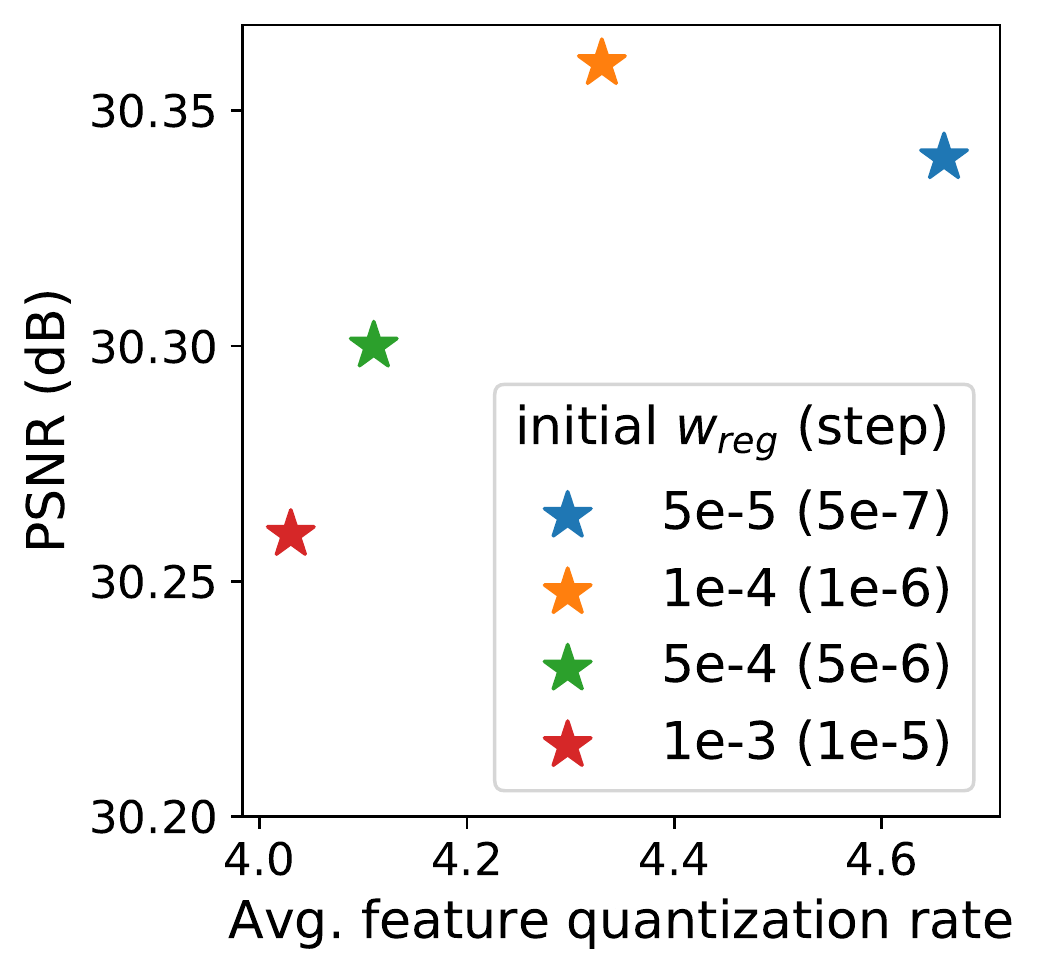}
    \caption{$w_{reg}$ of CARN-CADyQ}
    \label{fig:reb-wreg}
\end{figure}

$w_{reg}$ is the hyperparameter to control the trade-off between accuracy and efficiency.
We empirically chose the scalar value of $w_{reg}$ with the DIV2K validation PSNR, as in Fig.~\ref{fig:reb-wreg}.
A larger $w_{reg}$ gives further efficient network but at the cost of performance degradation.
Total BitOPs are manually controlled by $w_{reg}$, and how to allocate total BitOPs to each layer is controlled by $L_{wb}$ based on each layer's impact on overall performance.

\section{Implementation Details} \label{sec:sup-implementation}
Training is done with DIV2K dataset, which consists of eight hundred 2K-resolution images (index 0001-0800). 
Hundred images are selected (index 0801-0900) from the DIV2K validation set for validation.
The proposed framework is optimized using ADAM optimizer~\cite{kingma2014adam} with $\beta_1~\Equal~0.9$, $\beta_2~\Equal~0.999$, and $\epsilon~\Equal~10^{-8}$.
The mini-batch size is $16$, and the input patch size during training is $48\times48$.
The other training settings, such as the learning rate schedule, number of iterations, and data augmentation strategies, follow the settings of each baseline model. 
The initial learning rate is set as $0.01$ for the learnable scale parameter $a$ while the learning rate for the bit selector is the same as the learning rate for the SR network.
Table~\ref{tab:sup-implementation} covers the detailed settings for the implementation of the model.
Specifically, each epoch consists of 100 iterations, and the bit loss weight is initialized and then increased with \textit{step} amount for every epoch.
Also, the training data is augmented with random horizontal flips and ${90}^\circ$ rotations, and all the images are pre-processed by subtracting the mean RGB value of the DIV2K dataset.
\begin{table*}[h!]
\centering
\caption{ \textbf{Implementation details of CADyQ applied SR networks}}
\resizebox{1.\linewidth}{!}{
    \begin{tabular}{@{}c|cccc}
        \toprule
        &IDN & EDSR-baseline & SRResNet & CARN\\
        & -CADyQ & -CADyQ & -CADyQ & -CADyQ\\
        \midrule
        Epochs & 300 & 300 & 300 & 600 \\
        Initial learning rate & $10^{-4}$ & $10^{-4}$ & $10^{-4}$ & $10^{-4}$ \\
        Learning rate scheduling ($\gamma$, decay) & 0.5 at 150 & 0.5 at 150 & 0.5 at 150 & 0.5 at 400\\
        Bit selector initial learning rate & $10^{-4}$ & $10^{-4}$ & $10^{-4}$ & $10^{-3}$ \\
        Bit selector learning rate scheduling ($\gamma$, decay) & 0.5 at 150 & 0.5 at 150 & 0.5 at 150 & 0.5 at 400\\
        Bit loss initial weight & $10^{-4}$ & $10^{-4}$ & $10^{-4}$ & $10^{-4}$ \\
        Bit loss weight step & $10^{-6}$ & $10^{-6}$ & $10^{-6}$ & $10^{-6}$ \\
        \bottomrule
    \end{tabular}
}
\label{tab:sup-implementation}
\end{table*}

\section{Complexity Analysis} \label{sec:sup-complexity}
\subsection{Additional Complexity Analysis}
In addition to the CARN backbone models analyzed in the main manuscript, this section further analyzes the computational complexity of our framework applied to various SR networks such as IDN, EDSR-baseline, and SRResNet.
The computational complexity is measured with BitOPs, which denotes the number of operations weighted with the bit-width of the operands, and estimated energy.
Conventionally, BitOPs of a quantized convolution layer with weight $\bm{w}\in\mathbb{R}^{C\times C_{out}\times F\times F}$ of $b_w$-bit and input feature $\bm{x}\in\mathbb{R}^{N\times C\times H\times W}$ of $b$-bit, is calculated as 
$\frac{b_w}{32}\cdot\frac{b}{32}\cdot{2C}{C_{out}}F^2 NHW$.
However, since the bit-width of the feature varies for each patch in our framework, the BitOPs of a quantized convolution layer is calculated as 
$\sum_{i}^N{\frac{b_w}{32}\cdot\frac{b_i}{32}\cdot{2C}{C_{out}}F^2 NHW}$, where $b_i$ denotes the bit-width of i-th patch feature.
As stated in Section~\refnumber{4.5} of the main manuscript, our framework can process either the full input test image at once or process smaller patches in parallel, which are combined to construct the full image.
For both cases, our framework effectively reduces the computational resources while preserving the performance, as shown in Table~\ref{tab:sup-complexity}.

\begin{table*}[h!]
    \centering
    \caption{\textbf{Complexity analysis.}
    Computational costs of image-wise and patch-wise inference are respectively analyzed for models with IDN, EDSR-baseline, SRResNet backbone.
    Computational complexity is measured w.r.t. BitOPs of the feature extraction stage required for generating a $720p$ ($1820\times720$) image. 
    PSNR is measured on Urban100
    }
    \scalebox{0.99}{
        \begin{tabular}{lc rr rr}
            \toprule
            \multirow{3}{*}{Model} & \multirow{3}{*}{Params.} &\multicolumn{4}{c}{Inference Patch Size} \\
            \cmidrule(lr){3-6}
            && \multicolumn{2}{c}{Full Image} & \multicolumn{2}{c}{96$\times$96} \\
            \cmidrule(lr){3-4}\cmidrule(lr){5-6}
            && PSNR$_\uparrow$ & BitOPs$_\downarrow$ & PSNR$_\uparrow$ & BitOPs$_\downarrow$ \\
            \midrule
            IDN & 590.9K & 25.42 & 59.12G & 25.42 & 60.23G \\
            IDN-PAMS & 590.9K & 25.56 & 3.70G & 25.56 & 3.76G  \\
            \rowcolor{orange!25}
            IDN-CADyQ & 594.9K & 25.66 & 2.70G & 25.65 & 2.70G \\
            \midrule
            EDSR-baseline & 1517.6K & 26.04 & 135.90G & 26.04 & 138.45G \\
            EDSR-baseline-PAMS & 1517.6K & 25.94 & 8.49G & 25.94 & 8.65G \\
            \rowcolor{orange!25}
            EDSR-baseline-CADyQ & 1520.9K & 25.94 & 6.55G & 25.94 & 6.59G \\
            \midrule
            SRResNet & 1546.8K & 25.74 & 136.13G & 25.74 & 138.68G \\
            SRResNet-PAMS & 1546.8K & 25.85 & 8.51G & 25.85 & 8.67G \\
            \rowcolor{orange!25}
            SRResNet-CADyQ & 1555.4K & 25.92 & 6.22G & 25.92 & 6.21G \\
            \bottomrule
        \end{tabular}
    }
\label{tab:sup-complexity}
\end{table*}

\subsection{Overhead Analysis}
There exists computational overhead of our framework, which is from 1) overlapping patch-wise inference and 2) additional bit selector.
Generally, the overhead of patch-wise inference can be reduced with parallel processors as~\citenumber{28} and the overhead of overlapping patches is minimized by using small overlap regions (6 pixels from the boundary, in our case).
Nevertheless, as shown in Table~\ref{tab:reb-overhead}, overheads of patch-wise inference and bit selector are non-trivial.
However, CADyQ manages to overcome these overheads and successfully reduces latency/BitOPs, owing to its dynamic bit allocation.

\begin{table}[!h]
\centering
\caption{\textbf{Overhead analysis} for 4K image on CARN}
\resizebox{1.\linewidth}{!}{
\begin{threeparttable}
    \begin{tabular}{@{}l | ccc | ll}
        \toprule
        &Inference & Bit &\; \multirow{2}{*}{FQR} \;& Latency  & BitOPs \\
        &Type & Selector & & (diff w/ (a)) & (diff w/ (a))\\
        \midrule
        (a) PAMS &Image &\xm & 8.0 & 216.0 ms & 43.0 G \\
        (b) PAMS (patch) &Patch &\xm & 8.0 & 230.8 ms (\textcolor{red}{+14.8}) & 43.8 G (\textcolor{red}{+0.8})\\
        (c) PAMS (patch\Plus bit selector)\tnote{\textdagger} \; &Patch &\cm & 8.0 & 237.3 ms (\textcolor{red}{+20.9}) & 45.7 G (\textcolor{red}{+2.7})\\
        \midrule
        (d) CADyQ &Patch &\cm & 4.5 & 202.9 ms (\textcolor{blue}{-13.4}) & 29.2 G (\textcolor{blue}{-13.8})\\
        \bottomrule
    \end{tabular}
\begin{tablenotes}
    \item[\textdagger] (c) includes bit selector module, but unlike CADyQ, bit selector is forced to select 8-bit.
\end{tablenotes}
\end{threeparttable}
}
\label{tab:reb-overhead}
\end{table}

\clearpage

\section{Qualitative Results} \label{sec:sup-qualitative}
\begin{figure*}[!h]
\begin{center}\centering
\setlength{\tabcolsep}{0.05cm}
\newcommand{\h}{0.18\linewidth}
\newcommand{\hh}{0.36\linewidth}
\begin{tabular}{ccccc}
\multicolumn{2}{c}{ \scriptsize img1341 (Test4K)} & \multicolumn{2}{c}{\scriptsize Bit map (IDN-CADyQ)}\\
\multicolumn{2}{c}{\includegraphics[clip, trim=1cm 0cm 30cm 0cm, width=\hh]{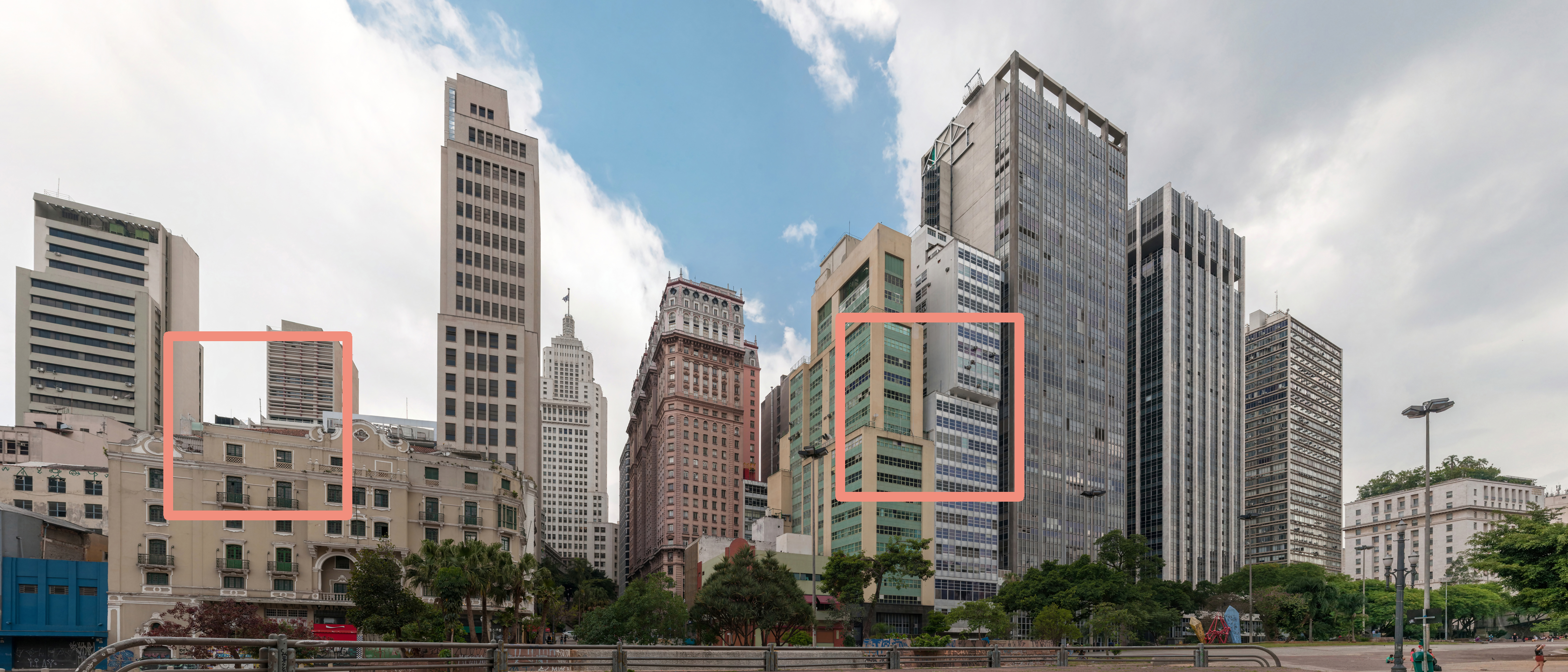} }&
\multicolumn{2}{c}{\includegraphics[clip, trim=1cm 0cm 30cm 0cm, width=\hh]{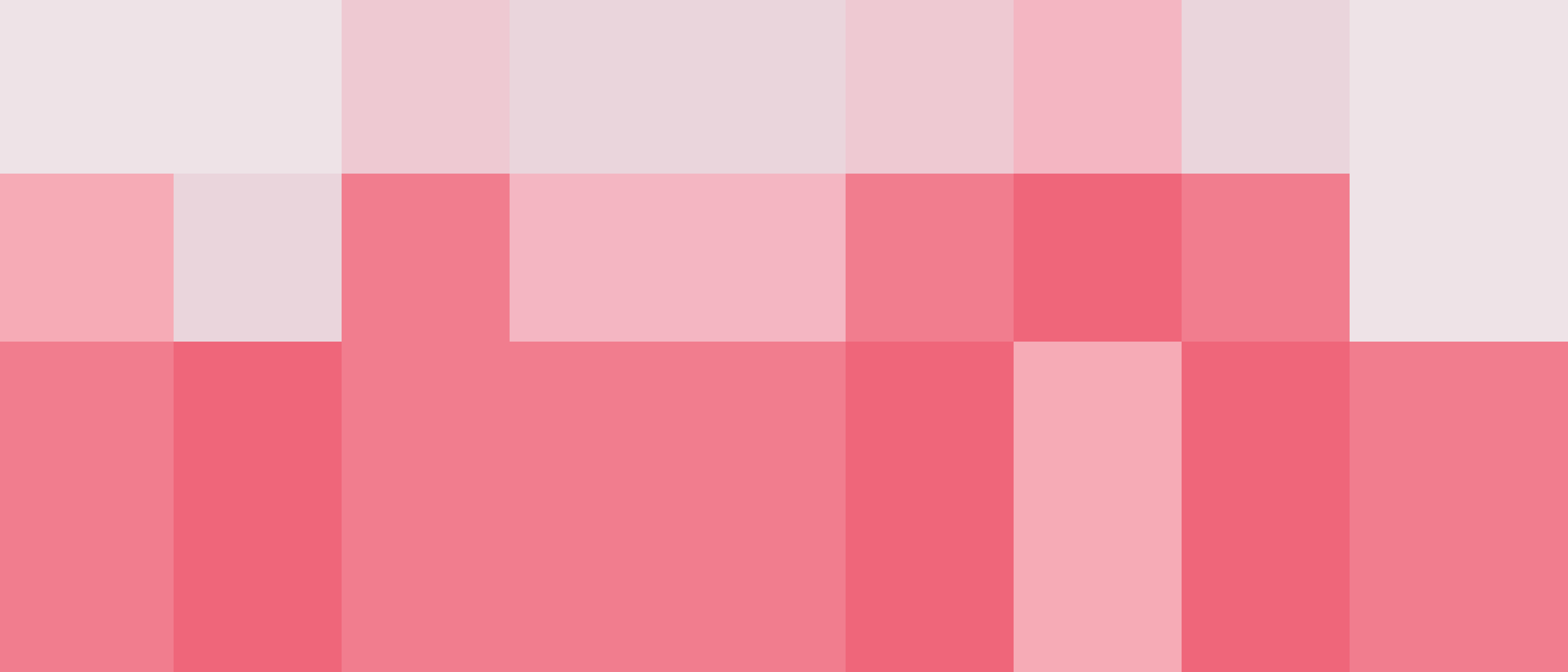} }\\
\scriptsize GT & \scriptsize IDN &\scriptsize IDN-PAMS & \scriptsize IDN-DAQ & \scriptsize IDN-CADyQ \\
\includegraphics[height=\h]{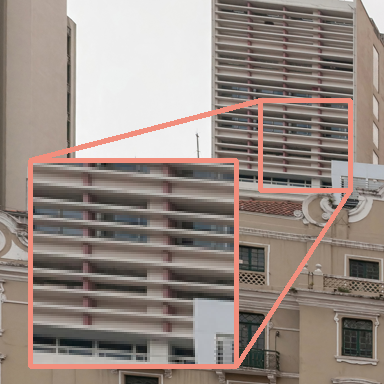} &
\includegraphics[height=\h]{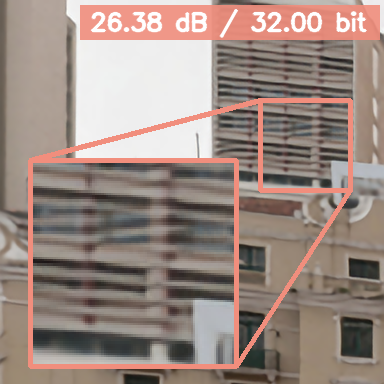} &
\includegraphics[height=\h]{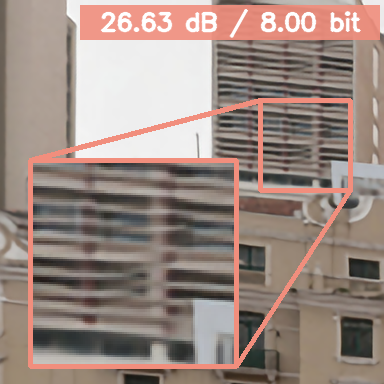} &
\includegraphics[height=\h]{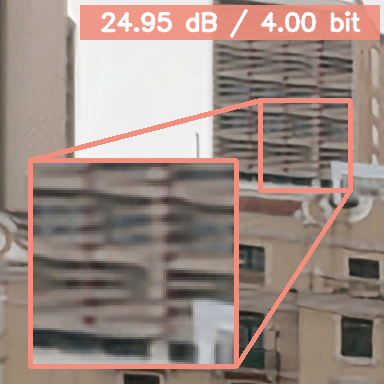} &
\includegraphics[height=\h]{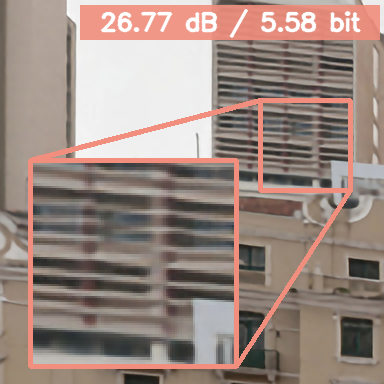} \\
\includegraphics[height=\h]{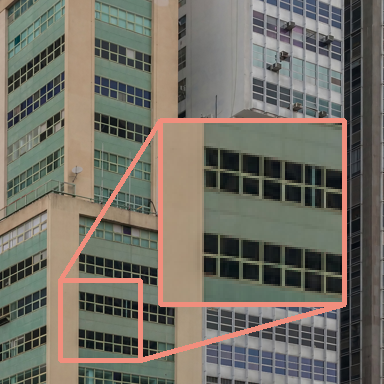} &
\includegraphics[height=\h]{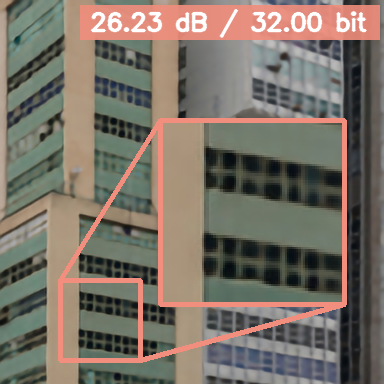} &
\includegraphics[height=\h]{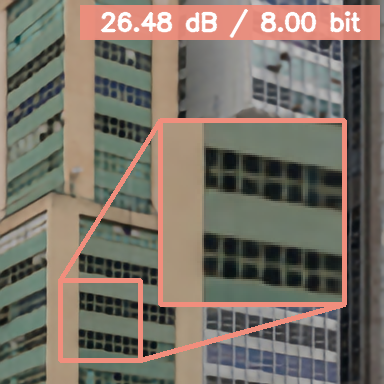} &
\includegraphics[height=\h]{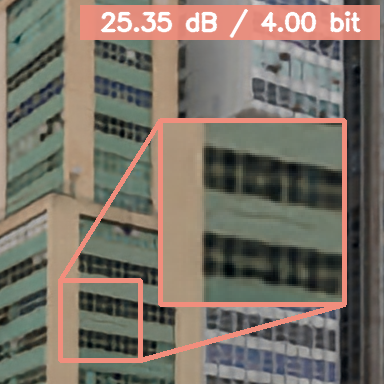} &
\includegraphics[height=\h]{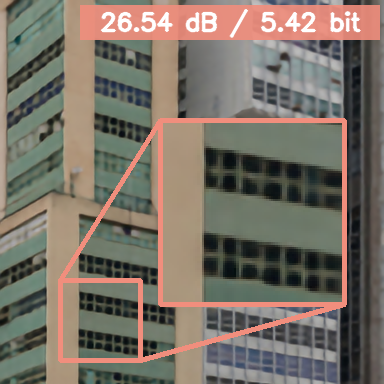} \\
\\
\multicolumn{2}{c}{ \scriptsize img1257 (Test4K)} & \multicolumn{2}{c}{\scriptsize Bit map (EDSR-baseline-CADyQ)}\\
\multicolumn{2}{c}{\includegraphics[clip, trim=0cm 5cm 5cm 0cm, width=\hh]{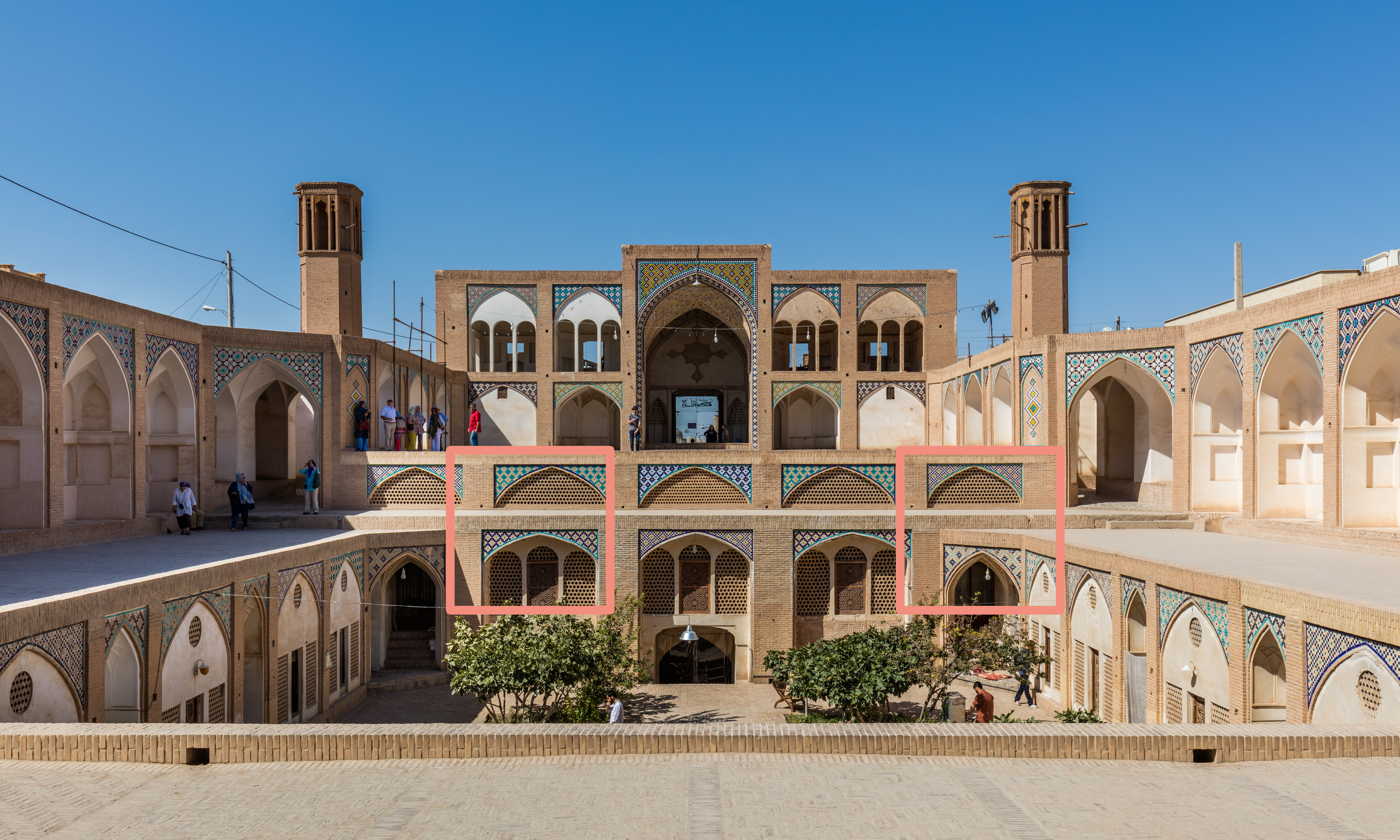} }&
\multicolumn{2}{c}{\includegraphics[clip, trim=0cm 5cm 5cm 0cm, width=\hh]{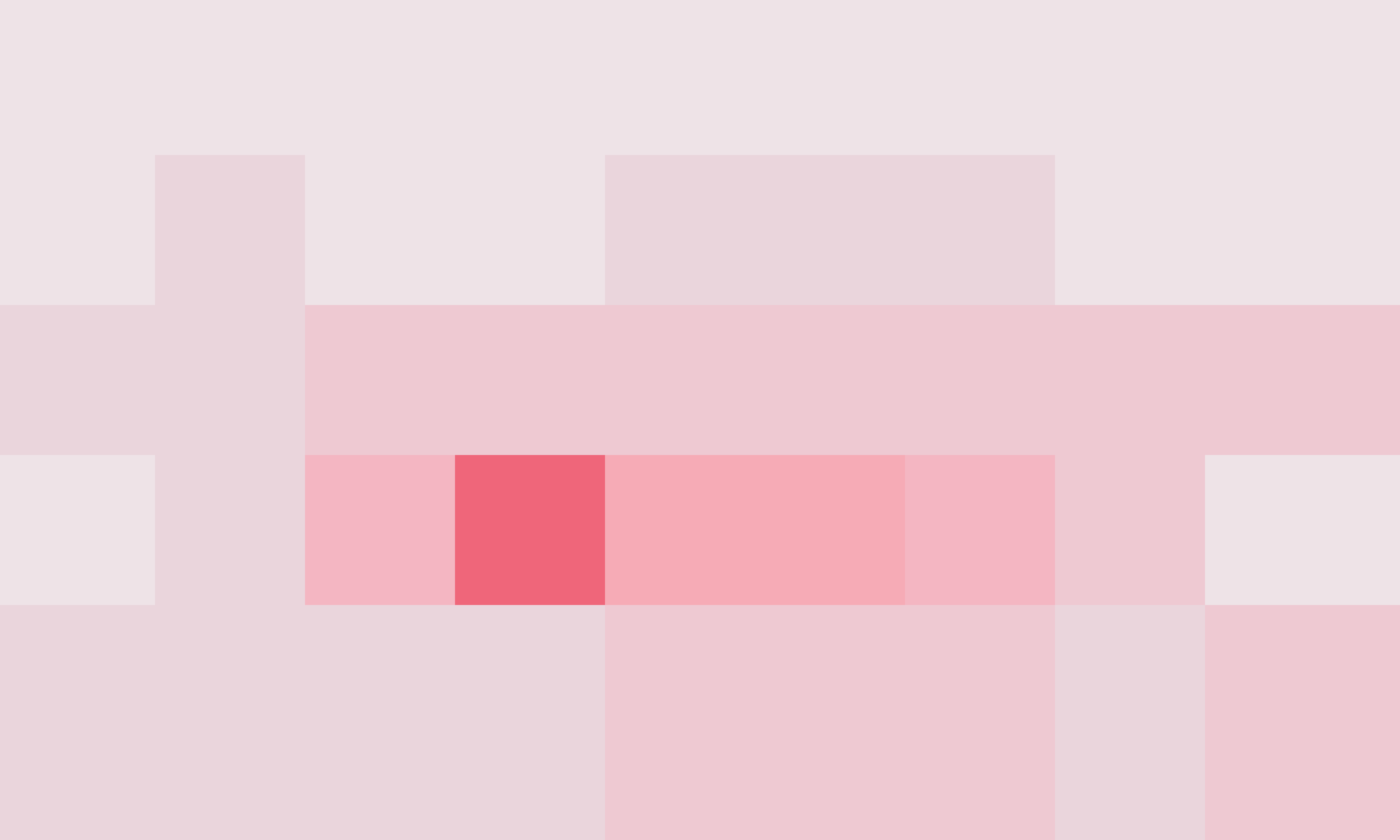}}\\
\scriptsize GT & \scriptsize EDSR-baseline &\scriptsize EDSR-baseline &\scriptsize EDSR-baseline & \scriptsize EDSR-baseline \\
&  & \scriptsize -PAMS & \scriptsize -DAQ & \scriptsize -CADyQ \\
\includegraphics[height=\h]{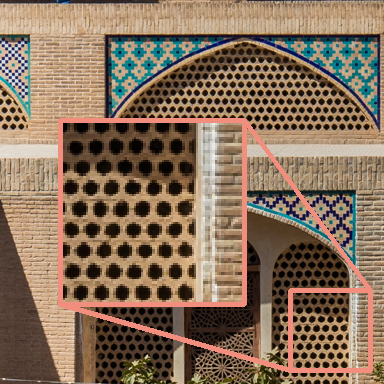} &
\includegraphics[height=\h]{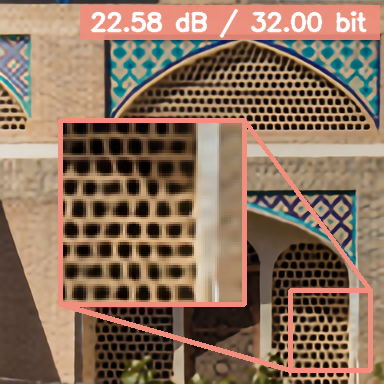} &
\includegraphics[height=\h]{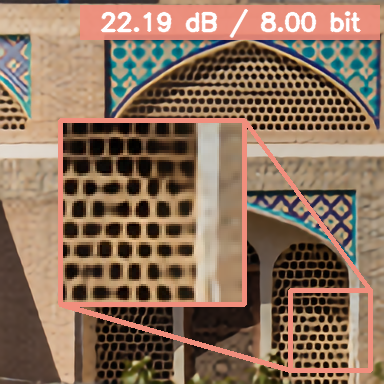} &
\includegraphics[height=\h]{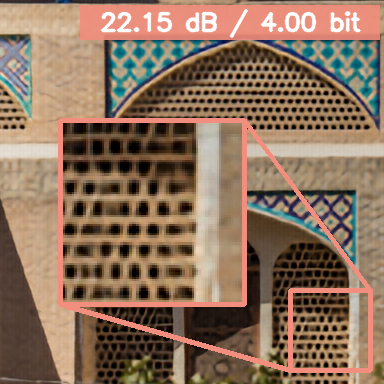} &
\includegraphics[height=\h]{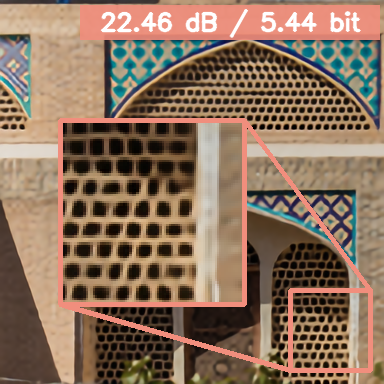} \\
\includegraphics[height=\h]{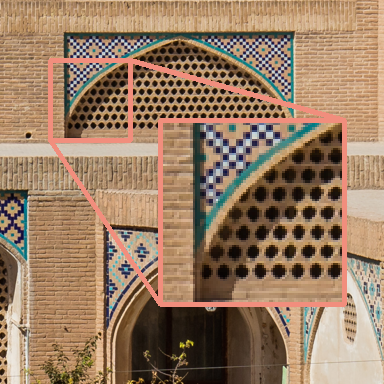} &
\includegraphics[height=\h]{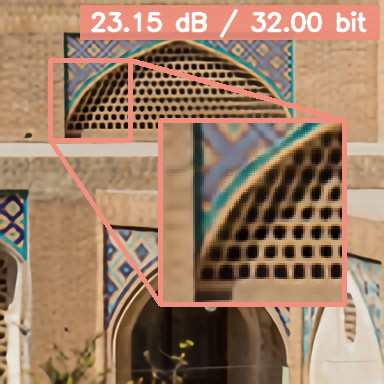} &
\includegraphics[height=\h]{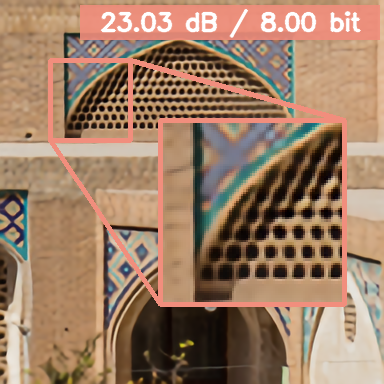} &
\includegraphics[height=\h]{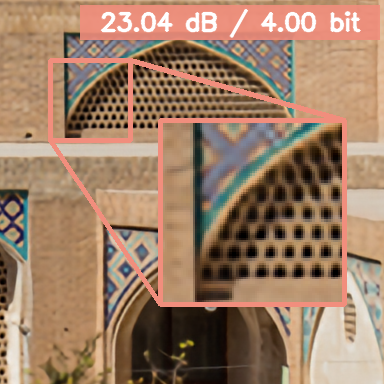} &
\includegraphics[height=\h]{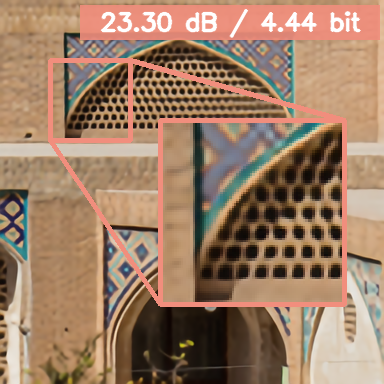} \\
\end{tabular}
\end{center}
\end{figure*}

\begin{figure*}[!t]
\centering
\setlength{\tabcolsep}{0.05cm}
\newcommand{\h}{0.18\linewidth}
\newcommand{\hh}{0.36\linewidth}
\begin{tabular}{ccccc}
\multicolumn{2}{c}{ \scriptsize img1339 (Test4K)} & \multicolumn{2}{c}{\scriptsize Bit map (SRResNet-CADyQ)}\\
\multicolumn{2}{c}{\includegraphics[clip, trim=0cm 0cm 5cm 0cm, width=\hh]{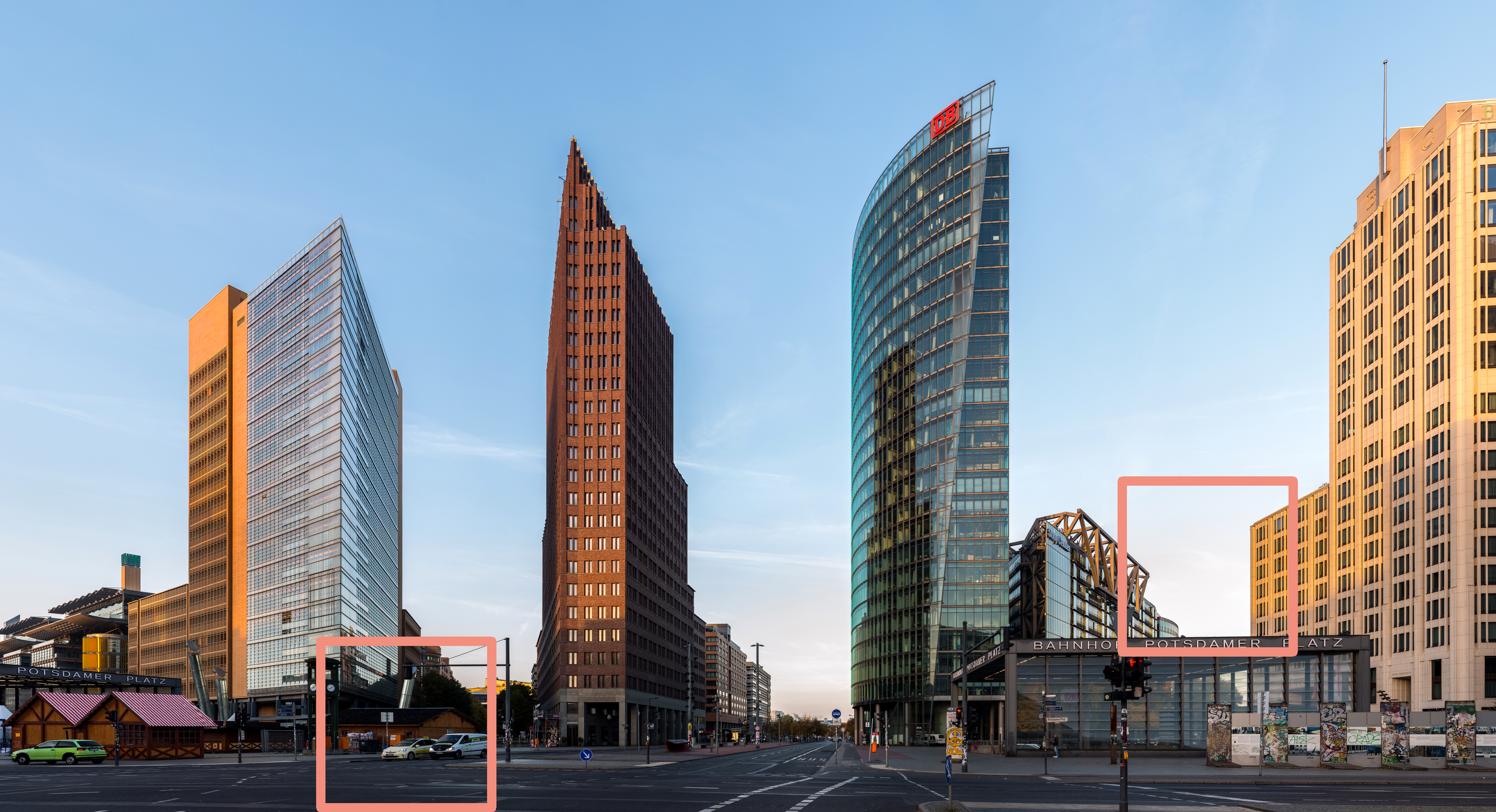} }&
\multicolumn{2}{c}{\includegraphics[clip, trim=0cm 0cm 5cm 0cm,width=\hh]{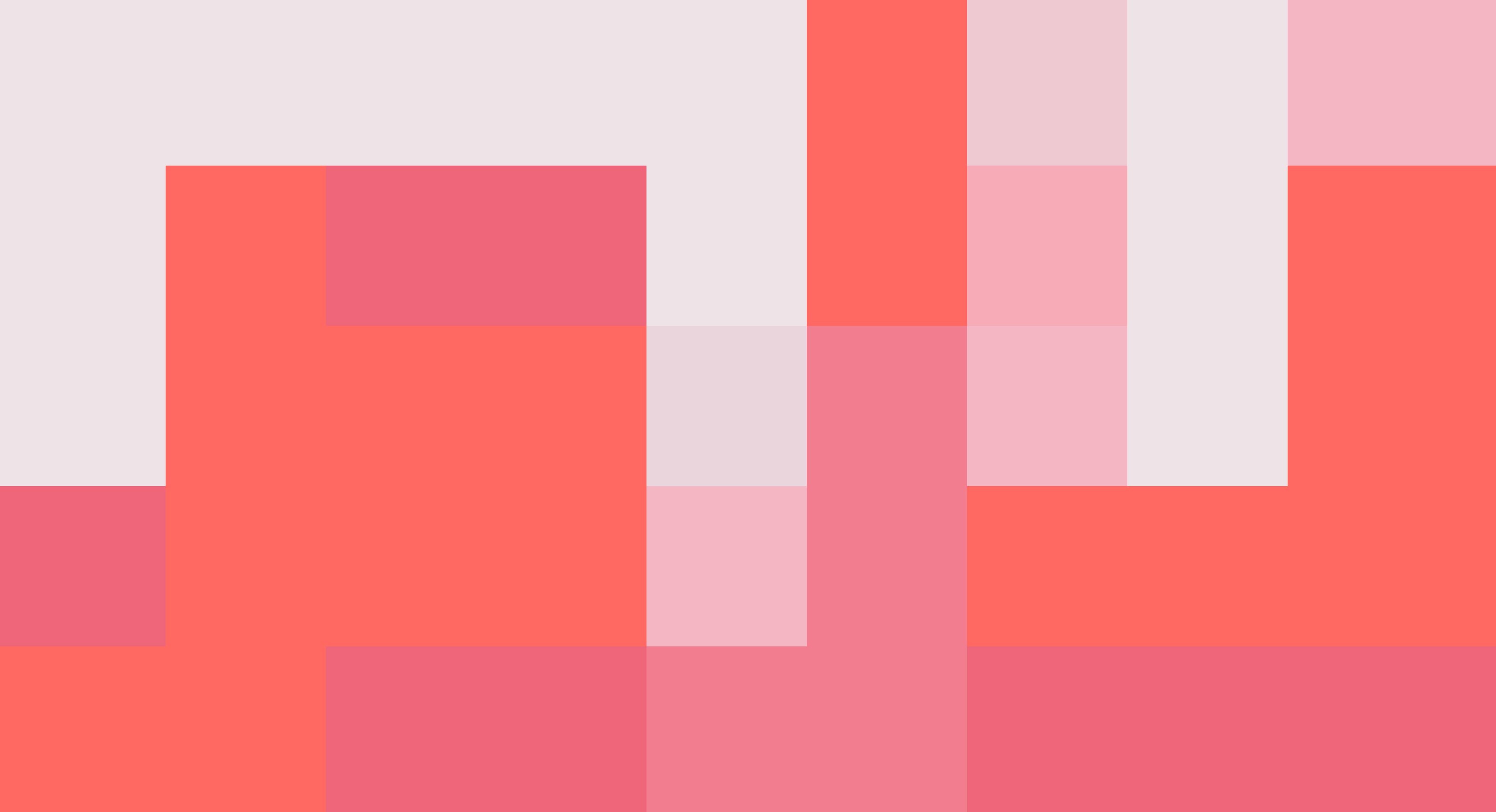}}\\
\scriptsize GT & \scriptsize SRResNet &\scriptsize SRResNet &\scriptsize SRResNet & \scriptsize SRResNet \\
&  & \scriptsize -PAMS & \scriptsize -DAQ & \scriptsize -CADyQ \\
\includegraphics[height=\h]{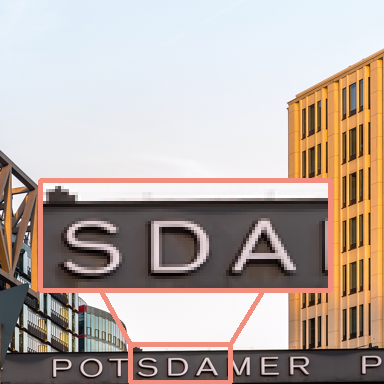} &
\includegraphics[height=\h]{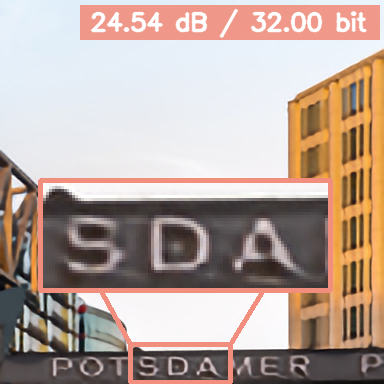} &
\includegraphics[height=\h]{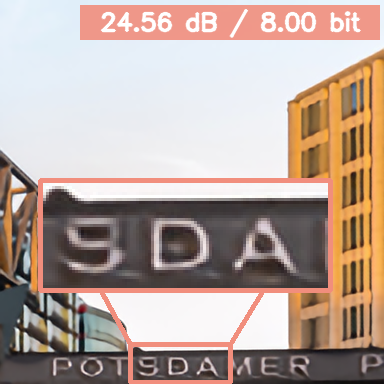} &
\includegraphics[height=\h]{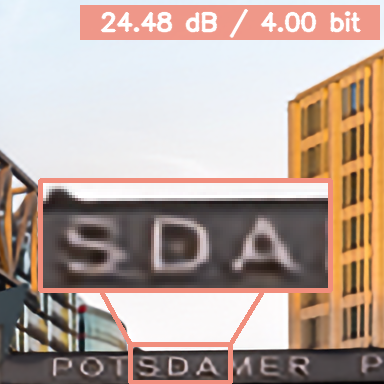} &
\includegraphics[height=\h]{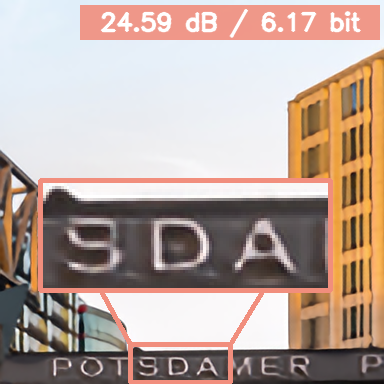} \\
\includegraphics[height=\h]{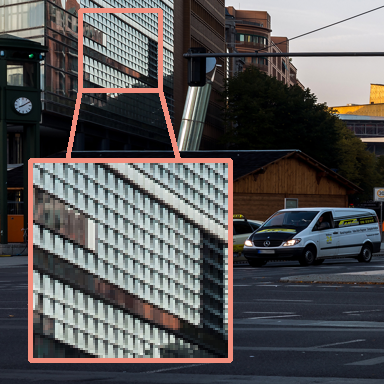} &
\includegraphics[height=\h]{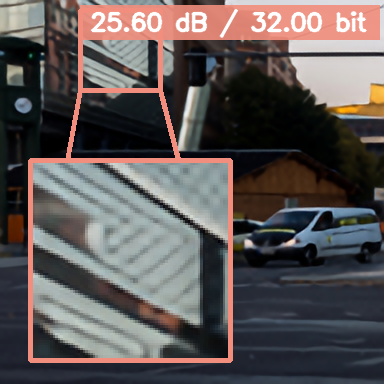} &
\includegraphics[height=\h]{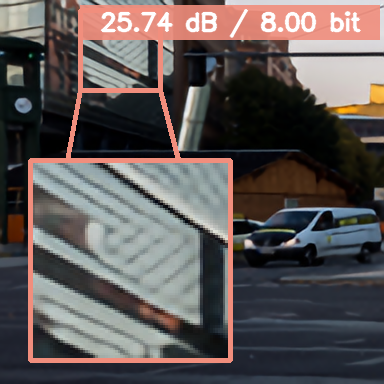} &
\includegraphics[height=\h]{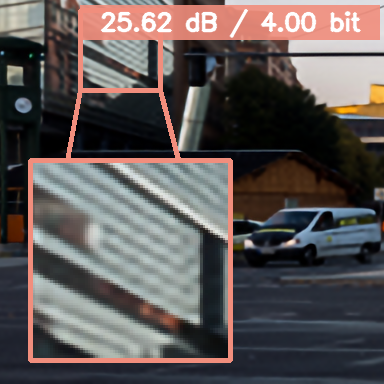} &
\includegraphics[height=\h]{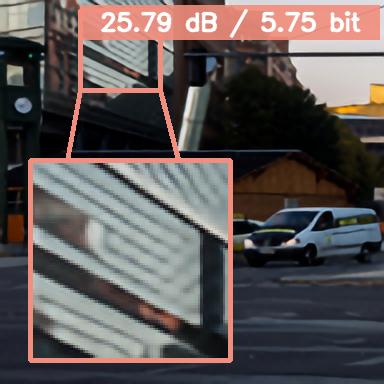} \\
\end{tabular}
\captionof{figure}{
\textbf{Qualitative results} from models of IDN, EDSR-baseline, and SRResNet backbone
}
\label{fig:sup-qualitative}
\end{figure*}

\section*{License of the Used Assets}

\begin{compactitem}[$\bullet$]
    \item DIV2K dataset~\citenumber{1} is made available for academic research purposes.
    \item Urban100 dataset~\citenumber{19} is made available at \url{https://github.com/jbhuang0604/SelfExSR}
    \item Test2K and Test4K dataset~\citenumber{28} is made available at \url{https://github.com/Xiangtaokong/ClassSR}
\end{compactitem}

\bibliographystyle{splncs04}
\bibliography{egbib}